\def\paperID{0000}
\def\confName{ICCV}
\def\confYear{2025}

\def\paperTitle{%
  \raisebox{-.01\height}{\includegraphics[scale=.06]{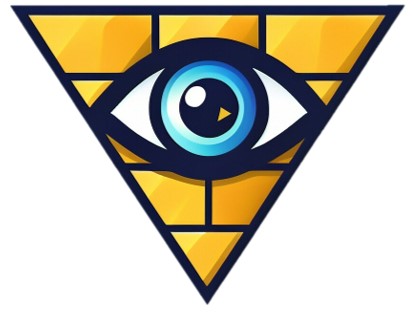}}%
  SpiritSight Agent: Advanced GUI Agent with One Look%
}

\def\authorBlock{
    Zhiyuan Huang$^{1*}$\qquad
    Ziming Cheng$^{1,2}$\thanks{Equal contribution.}\qquad
    Junting Pan$^3$\qquad
    Zhaohui Hou$^1$\qquad
    Mingjie Zhan$^1$\\
    \vspace{1em}
    $^1$SenseTime Research,
    $^2$Beijing University of Posts and Telecommunications,
    $^3$MMLab, CUHK  \\
}

\newif\ifreview 
\newif\ifarxiv \newcommand{\arxiv}{\arxivtrue}
\newif\ifcamera 
\newif\ifrebuttal 

\arxiv %

\pdfoutput=1
\documentclass[10pt,twocolumn,letterpaper]{article}
\ifreview \usepackage[review]{cvpr} \fi
\ifarxiv \usepackage[pagenumbers]{cvpr} \fi
\ifrebuttal \usepackage[rebuttal]{cvpr} \fi
\ifcamera \usepackage{cvpr} \fi

\usepackage{graphicx}	
\usepackage{amsmath}	
\usepackage{amssymb}	
\usepackage{booktabs}
\usepackage{times}
\usepackage{microtype}
\usepackage{epsfig}
\usepackage{caption}
\usepackage{float}
\usepackage{placeins}
\usepackage{color, colortbl, tcolorbox}
\usepackage{stfloats}
\usepackage{enumitem}
\usepackage{tabularx}
\usepackage{xstring}
\usepackage{multirow}
\usepackage{xspace}
\usepackage{url}
\usepackage{subcaption}
\usepackage{xcolor}
\usepackage[hang,flushmargin]{footmisc}

\ifcamera \usepackage[accsupp]{axessibility} \fi

\ifarxiv  \fi

\newcommand{\R}[1]{{%
    \textbf{%
        \ifstrequal{#1}{1}{\textcolor{red}{R#1}}{%
        \ifstrequal{#1}{2}{\textcolor{blue}{R#1}}{%
        \ifstrequal{#1}{3}{\textcolor{magenta}{R#1}}{%
        \ifstrequal{#1}{4}{\textcolor{teal}{R#1}}{%
                           \textcolor{cyan}{R#1}%
        }}}}%
    }%
}}

\usepackage{xr-hyper}

\makeatletter
\newcommand*{\addFileDependency}[1]{
  \typeout{(#1)}
  \@addtofilelist{#1}
  \IfFileExists{#1}{}{\typeout{No file #1.}}
}

\makeatother

\definecolor{cvprblue}{rgb}{0.21,0.49,0.74}
\usepackage[pagebackref,breaklinks,colorlinks,allcolors=cvprblue]{hyperref}
\usepackage[capitalize]{cleveref}
\crefname{section}{Sec.}{Secs.}
\crefname{table}{Table}{Tables}
\crefname{figure}{Fig.}{Figs.}

\ifarxiv \crefname{appendix}{App.}{Apps.}
\else \crefname{appendix}{Suppl.}{Suppls.} \fi

\renewcommand{\arraystretch}{1.15}

\begin{document}
\title{\paperTitle}
\author{\authorBlock}
\maketitle

\begin{abstract}
Graphical User Interface (GUI) agents show amazing abilities in assisting human-computer interaction, automating human user's navigation on digital devices. An ideal GUI agent is expected to achieve high accuracy, low latency, and compatibility for different GUI platforms. Recent vision-based approaches have shown promise by leveraging advanced Vision Language Models (VLMs). While they generally meet the requirements of compatibility and low latency, these vision-based GUI agents tend to have low accuracy due to their limitations in element grounding. To address this issue, we propose \textbf{SpiritSight}, a vision-based, end-to-end GUI agent that excels in GUI navigation tasks across various GUI platforms. First, we create a multi-level, large-scale, high-quality GUI dataset called \textbf{GUI-Lasagne} using scalable methods, empowering SpiritSight with robust GUI understanding and grounding capabilities. Second, we introduce the \textbf{Universal Block Parsing (UBP)} method to resolve the ambiguity problem in dynamic high-resolution of visual inputs, further enhancing SpiritSight's ability to ground GUI objects. Through these efforts, SpiritSight agent outperforms other advanced methods on diverse GUI benchmarks, demonstrating its superior capability and compatibility in GUI navigation tasks. Models and datasets are available at \url{https://hzhiyuan.github.io/SpiritSight-Agent}.
\end{abstract}

\begin{figure}[t]
  \centering
    \includegraphics[width=0.95\linewidth]{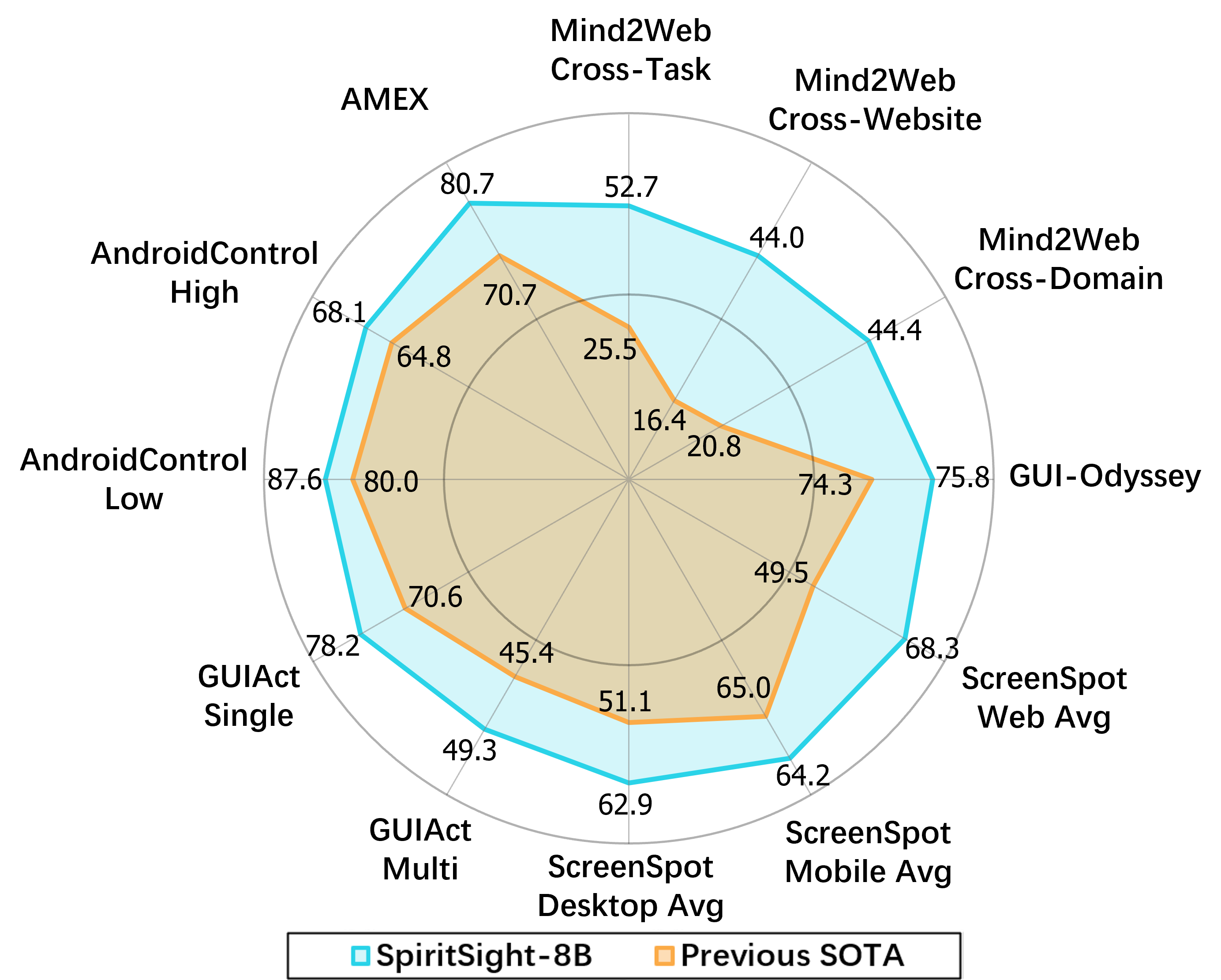}
    \caption{Our SpiritSight agent achieves new state-of-the-art (SOTA) performance across various benchmarks in web, mobile, and desktop scenarios.}
  \label{fig:radar}
\end{figure}

\section{Introduction}
\label{sec:intro}

Graphical User Interface (GUI) automation has long been pursued by people along with the development of the modern digital devices. Thanks to recent advances in Large Language Models (LLMs), GUI agents are constructed to assist users in interacting with graphical interfaces, automatically making action decisions based on observations of environment and user's instruction. 

Current approaches can be divided into three categories based on their input modalities. Language-based and vision-based approaches use Hyper Text Markup Language (HTML) or Extensible Markup Language (XML) and screenshots as input~\cite{zheng2023synapse, huq2023claud2-agent, deng2024mind2web,wan2024omniparser, lai2024autowebglm, lee2024gist, yin2024agent-lumos, hong2024cogagent, cheng2024seeclick, chen2024guicourse}, respectively, while vision-language-based approaches combine both HTML and screenshots as input~\cite{furuta2023webgum, thil2024CC-Net, kil2024dualVCR, zheng2024seeact}.

The language-based and vision-language-based methods typically applied only in the web domain, and often limited by the excessive length of HTML or security concerns~\cite{zhan2024injecagent, wu2024adversarial, liao2024eia} regarding it. The vision-based approaches exhibit compatibility across various GUI platforms, as acquiring screenshots is generally easier than obtaining hierarchical data from non-web platforms.

However, vision-based approaches struggle to ground GUI elements (\eg buttons, text boxes) from the visual input~\cite{zheng2024seeact}. Some works solve this problem by introducing additional tools, such as optical character recognition (OCR) and icon recognition models, to convert the visual grounding task into a language QA task~\cite{wang2024Mobile-agent, wang2024Mobile-Agent-v2}. This may increase the complexity and inference latency of the agent system. Others attempt to collect large-scale training data through manual synthesis~\cite{shi2017MiniWoB, liu2018MiniWoB++, lee2023pix2struct} or human annotation~\cite{yao2022webshop, deng2024mind2web, rawles2024AndroidintheWild, chen2024GUI-WORLD, chai2024amex, lu2024GUI-Odyssey, lu2024weblinx}, while these data are either unrealistic or costly. Additionally, recent advanced Vision Language Models (VLMs)~\cite{ye2023ureader, chen2024internvl2} adopt dynamic high-resolution~\cite{ye2023ureader, chen2024internvl2} strategy to better suit high-resolution inputs, which we find may introduce ambiguity to the model learning process in GUI scenarios.

To address the aforementioned challenges, we proposed an end-to-end, vision-based GUI agent——SpiritSight, which has strong ability in GUI navigation task. Our contributions are summarized as follows:

\textbf{Firstly, we propose GUI-Lasagne, a multi-level, large-scale and high-quality GUI dataset composed of 5.73 million samples to enhance our model's GUI understanding and grounding capabilities.} The dataset is collected from the real-world and filtered through carefully designed rules to ensure data quality. It is also constructed hierarchically and consists of three levels of components: text/icon recognition and grounding data, function grounding data, and GUI navigation data. The first two parts constitute 90\% of the total dataset and are collected for free, thus significantly reducing the data collection cost.

\textbf{Secondly, We introduce a Universal Block Parsing (UBP) method to resolve the ambiguity problem in dynamic high-resolution inputs.} With the effectiveness of UBP, SpiritSight gains an improved elements grounding capability and achieves significant performances in GUI navigation tasks.

\begin{figure}[t]
  \centering
    \includegraphics[width=0.95\linewidth]{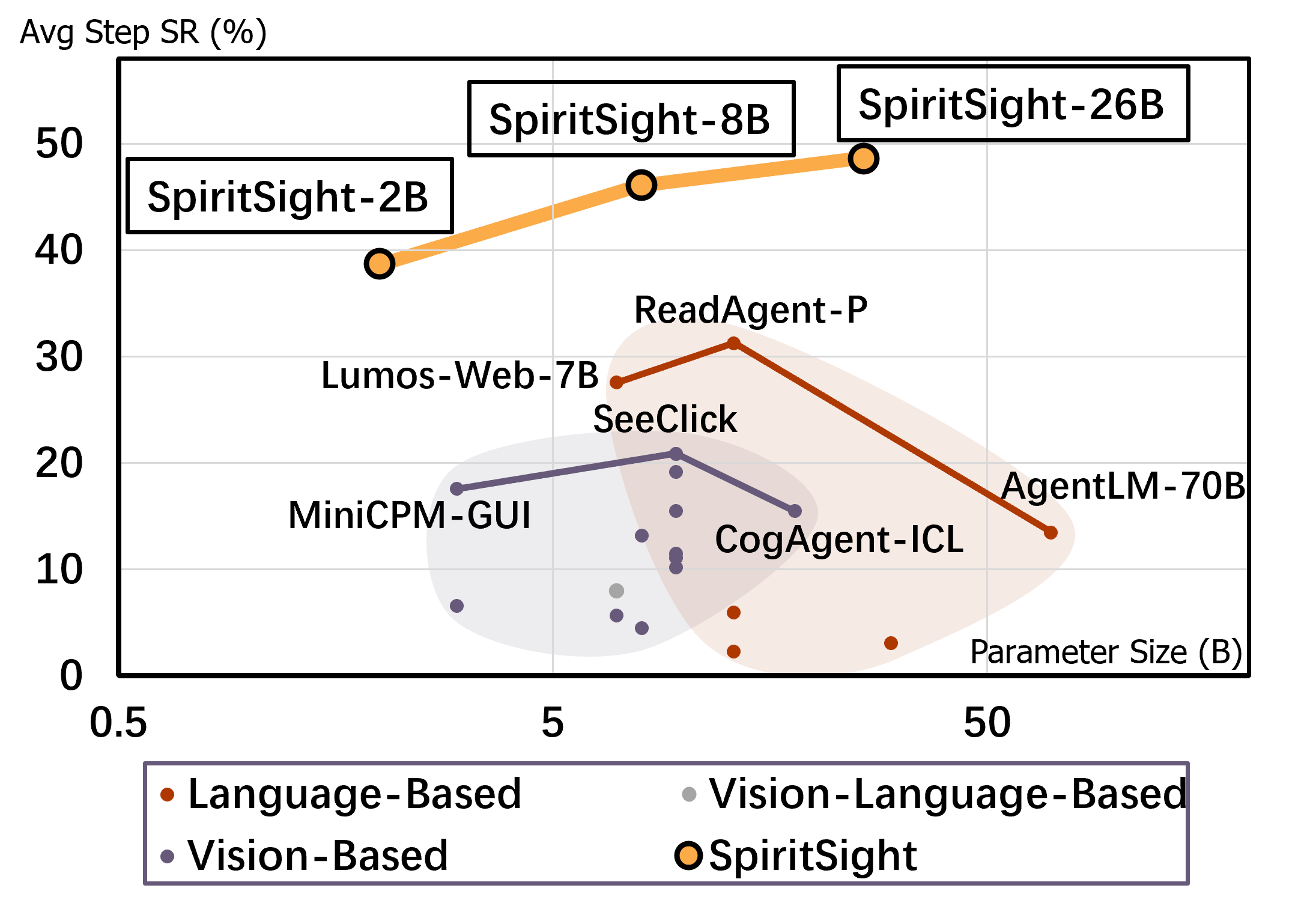}
    \caption{Comparison of the average step success rate on Multimodal-Mind2Web benchmark of our SpiritSight agent of three sizes (2B, 8B, 26B) with various previous methods.}
  \label{fig:mind2web-small}
\end{figure}

\begin{figure}[t]
  \centering
    \includegraphics[width=0.95\linewidth]{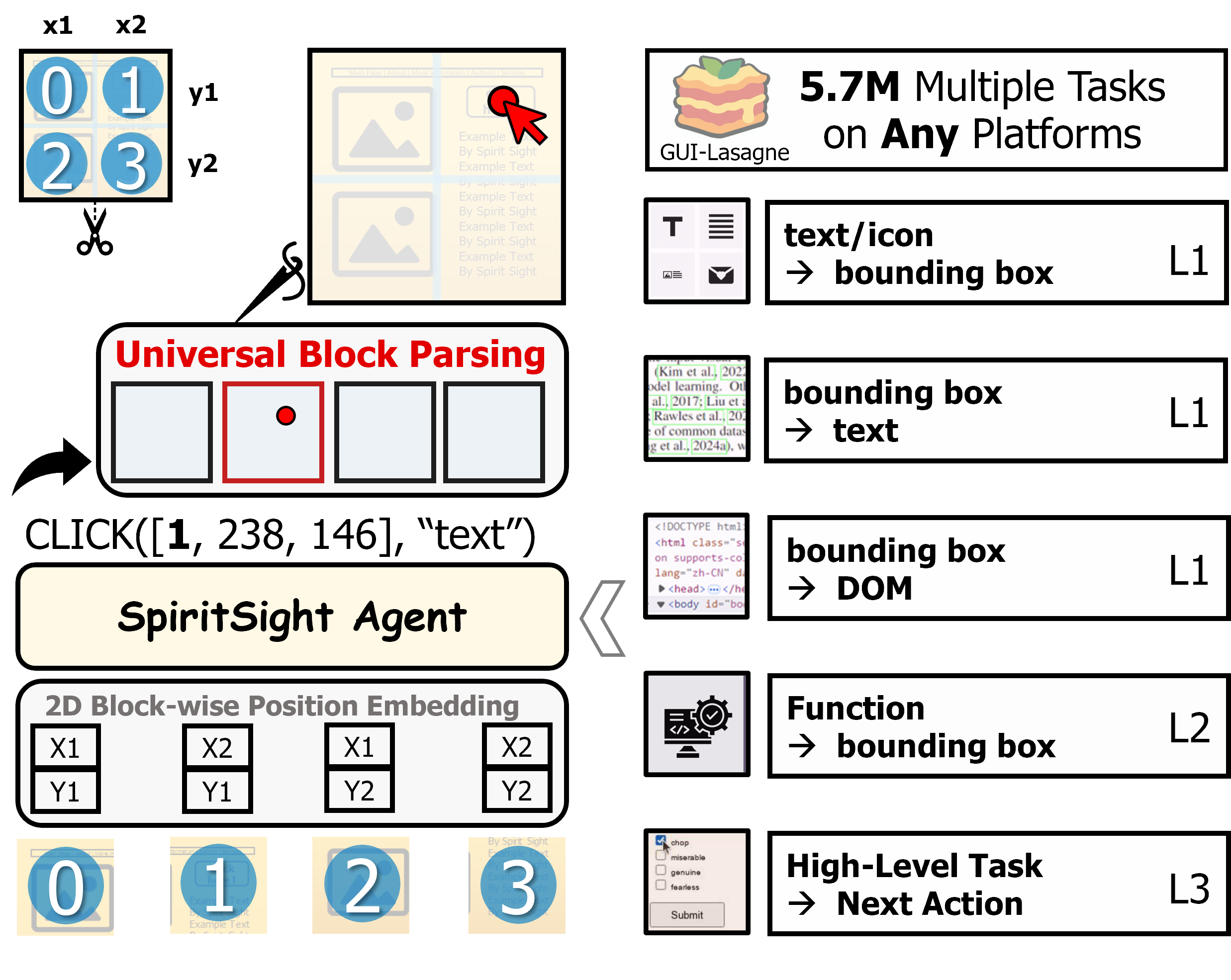}
    \caption{The overview of our SpiritSight agent. We develop a large-scale, multi-level, high-quality pre-training dataset that equips SpiritSight with three levels of comprehensive GUI knowledge. Additionally, we introduce a Universal Block Parsing (UBP) method to enhance SpiritSight's grounding capabilities.}
  \label{fig:overview}
\end{figure}

\textbf{Thirdly, we release and evaluate SpiritSight agent in various GUI benchmarks and it exhibits impressive performance among them.} SpiritSight agent is pre-trained on GUI-Lasagne dataset and employs UBP method to ground target elements, as shown in \cref{fig:overview}. We release three versions of SpiritSight with different model size: SpiritSight-26B, SpiritSight-8B, and SpiritSight-2B. The standard SpiritSight-8B consistently outperforms previous state-of-the-art methods across various GUI benchmarks, as shown in \cref{fig:radar}.
On the Multimodal-Mind2Web~\cite{deng2024mind2web} benchmark, the SpiritSight series outperforms all kinds of methods—including language-based, vision-based and even vision-language-based methods—in the non-candidate setting, as shown in \cref{fig:mind2web-small}.

\begin{table*}[t]
\begin{center}
\small
\begin{tabular}
{c@{\hspace{4pt}}c@{\hspace{4pt}}c@{\hspace{4pt}}c@{\hspace{4pt}}c@{\hspace{6pt}}c@{\hspace{6pt}}c@{\hspace{6pt}}c@{\hspace{6pt}}c@{\hspace{6pt}}c@{\hspace{6pt}}c@{\hspace{6pt}}c@{\hspace{6pt}}c}
\hline
\multirow{2}{*}{\textbf{}} &
  \multirow{2}{*}{\begin{tabular}[c]{@{}c@{}}Model\\ Size\end{tabular}} &
  \multirow{2}{*}{\begin{tabular}[c]{@{}c@{}}Input\\ Modality\end{tabular}} &
  \multirow{2}{*}{\begin{tabular}[c]{@{}c@{}}Select \\From Top\end{tabular}} &
  \multicolumn{3}{c}{Cross-Task} &
  \multicolumn{3}{c}{Cross-Website} &
  \multicolumn{3}{c}{Cross-Domain} \\ \cline{5-13} 
                    &       &       &       & Ele.Acc & Op.F1 & Step SR & Ele.Acc & Op.F1 & Step SR & Ele.Acc & Op.F1 & Step SR \\ \hline\hline
AutoWebGLM~\cite{lai2024autowebglm}         & 6B    & Text     & \checkmark       & -       & -     & 66.4\%    & -       & -     & 56.4\%    & -       & -     & 55.8\%    \\
LLaMA2-7B~\cite{lai2024autowebglm}          & 7B    & Text     & \checkmark       & -       & -     & 52.7\%    & -       & -     & 47.1\%    & -       & -     & 50.3\%    \\
CogAgent~\cite{hong2024cogagent}           & 18B   & Image     & \checkmark       & -       & -     & 62.3\%    & -       & -     & 54.0\%    & -       & -     & 59.4\%    \\
HTML-T5-XL~\cite{gur2023webagent}         & 3B    & Text     & \checkmark       & 76.4\%       & 78.8\%     & 71.5\%    & 68.4\%       & 71.0\%     & 62.2\%    & 73.0\%       & 75.6\%     & 67.1\%    \\
\hline\hline
SeeAct~\cite{zheng2024seeact}            & -     & Text+Image     & ×       & 46.4\%    & 73.4\%  & 40.2\%    & 38.0\%    & 67.8\%  & 32.4\%    & 42.4\%    & 69.3\%  & 36.8\%   \\
ReadAgent-P~\cite{lee2024gist}        & 340B  & Text     & ×       & 33.7\%    & 72.5\%  & 29.2\%    & 37.4\%    & 75.1\%  & 31.1\%    & 37.2\%    & 76.3\%  & 33.4\%    \\
 \hline
MiniCPM-GUI~\cite{chen2024guicourse}         & 3B    & Image     & ×       & 23.8\%    & 86.8\%  & 20.8\%    & 20.3\%    & 81.7\%  & 17.3\%    & 17.9\%    & 74.5\%  & 14.6\%    \\
Fuyu-GUI~\cite{fuyu-8b}            & 8B    & Image     & ×       & 19.1\%    & 86.1\%  & 15.6\%    & 13.9\%    & 80.7\%  & 12.2\%    & 14.2\%    & 83.1\%  & 11.7\%    \\
SeeClick~\cite{cheng2024seeclick}            & 9.6B  & Image     & ×       & 28.3\%    & 87.0\%  & 25.5\%    & 21.4\%    & 80.6\%  & 16.4\%    & 23.2\%    & 84.8\%  & 20.8\%    \\
OmniParser~\cite{wan2024omniparser} & - & Image & × & 42.4\% & 87.6\% & 39.4\% & 41.0\% & 84.8\% & 36.5\% & 45.5\% & 85.7\% & 42.0\% \\
\hline
SpiritSight-2B      & 2B    & Image     & ×       & 51.7\%    & 87.2\%  & 44.9\%    & 44.0\%    & 83.6\%  & 37.8\%    & 42.4\%    & 83.5\%  & 36.9\%    \\
SpiritSight-8B      & 8B    & Image     & ×       & 59.2\%    & 88.9\%  & 52.7\%    & 52.2\%    & 84.7\%  & 44.0\%    & 50.1\%    & 86.0\%  & 44.4\% \\
SpiritSight-26B     & 26B   & Image     & ×       & \textbf{60.5\%} & \textbf{89.7\%} & \textbf{54.7\%} & \textbf{57.0\%} & \textbf{85.7\%} & \textbf{48.1\%} & \textbf{54.1\%} & \textbf{87.2\%} & \textbf{49.2\%} \\ \hline
\end{tabular}
\end{center}
\caption{Comparison of SpiritSight agent with various previous methods on Multimodal-Mind2Web benchmark. SpiritSight significantly outperforms all methods that do not rely on any candidate element, including vision-based methods, language-based methods, and even vision-language-based methods.}
\label{sota-mind2web}
\end{table*}

\section{Related Works}
\label{sec:related_works}

\subsection{Language-based and vision-language-based GUI Agent}
Several works leverage the capabilities of large-scale language models (LLMs) to construct GUI agents. It is noticed that they are mostly multi-stage architectures. Mind2Web~\cite{deng2024mind2web} utilizes a lightweight language model to extract candidate elements from HTML, followed by a ranking model that sorts these elements based on task descriptions and historical actions. Finally, a large language model predicts the current actions and selects the target elements to operate on. WebAgent~\cite{gur2023webagent} first uses an encoder-decoder model to generate low-level instructions and relevant HTML code snippets, then uses another decoder to produce executable Python code. AutoWebGLM~\cite{lai2024autowebglm} simplifies HTML code through manually designed rules before predicting the executable action codes.

Other vision-language-based works leverage both GUI screenshots and hierarchical HTML/XML to enhance the robustness of GUI agents. WebGUM~\cite{furuta2023webgum} and CC-Net~\cite{thil2024CC-Net} use ResNet and ViT to extract features from screenshots, respectively. The image embeddings are then combined with text embeddings and fed into a multi-modal transformer. SeeAct~\cite{zheng2024seeact} and AppAgent~\cite{yang2023appagent} identify all interactive elements using HTML or XML data. They then assign each interactive element a unique identifier in the screenshot before feeding the screenshot into the model.

These language-based and vision-language-based methods, which rely on hierarchical information, exhibit several limitations: (1) Hierarchical representations, such as HTML or XML, are not consistently available across different platforms. And even this information is available, differences in their internal rules make language-based GUI agents less compatible; (2) HTML often contains redundant and customized information, requiring additional models or extensive manually crafted rules for effective filtering. (3) Text-based GUI agents are susceptible to injection attacks~\cite{zhan2024injecagent, wu2024adversarial, liao2024eia}, where malicious instructions hidden in HTML can easily lead to erroneous or unsafe actions.

\subsection{Vision-based GUI Agent}
Recently, some vision-based approaches have been proposed to overcome the drawbacks of language-based methods. Some of them~\cite{shaw2023pix2act, hong2024cogagent, cheng2024seeclick, baechler2024screenai} are single-stage methods that only use GUI screenshots as input for VLMs and output the next action in an end-to-end manner. However, these agents perform worse on GUI benchmarks compared to other methods. Two primary reasons may be the relatively low input resolution of their base models and the insufficient scale and quality of the training dataset. SeeAct~\cite{zheng2024seeact}, MobileAgent~\cite{wang2024Mobile-agent} and MobileAgent-v2~\cite{wang2024Mobile-Agent-v2} are two-stage methods. They use GPT-4V instead of open-source VLMs as base models and find that even the top models struggle with element grounding. Consequently, they introduce additional tools such as OCR and icon recognition models to assist with element grounding, which may increase the complexity and inference latency of the agent system. Overall, the current VLMs exhibit limited ability in handling GUI grounding tasks, thus constrains the navigation capabilities of single-stage vision-based GUI agents. We discuss additional related works on LLMs and VLMs in \cref{appendix:related-work}.

\begin{figure*}[t]
  \centering
    \includegraphics[width=\linewidth]{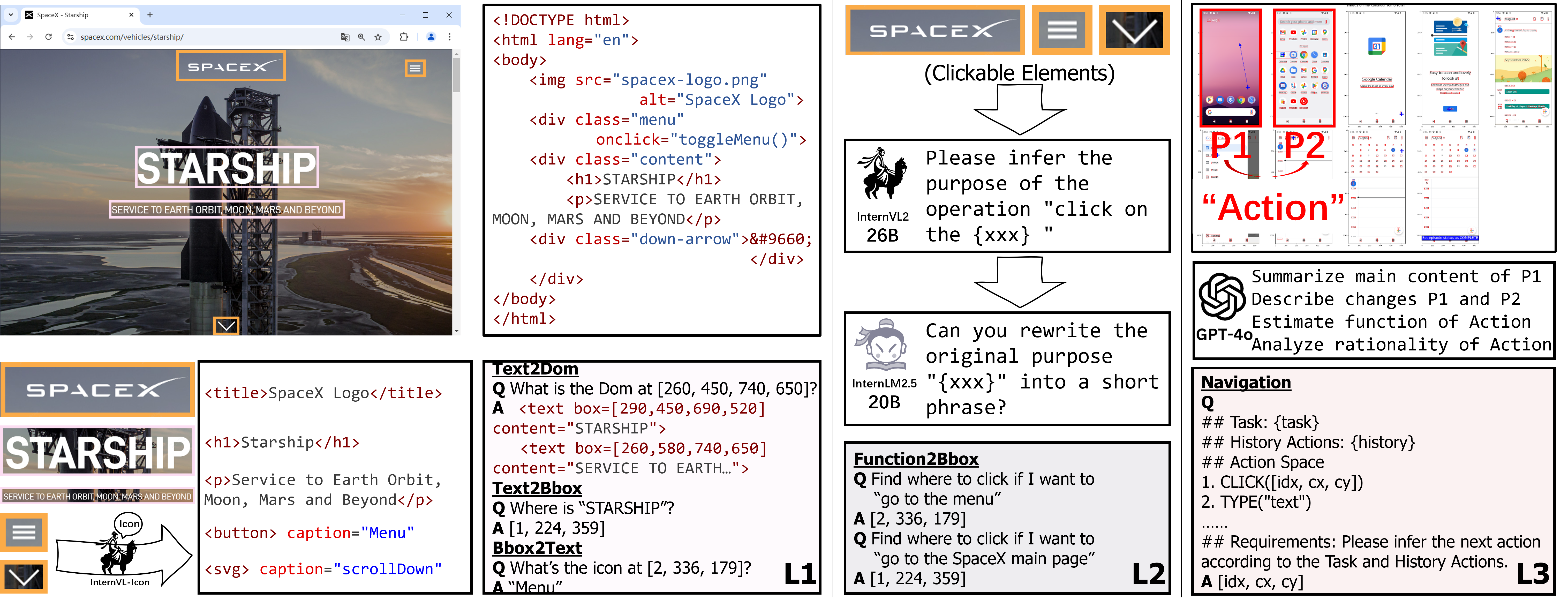}
    \caption{The collection pipeline of our GUILasagne dataset. The left, middle and right parts show the construction of level-1, level-2, and level-3 data, respectively.}
    \label{fig:data-construction}
\end{figure*}

\section{Data Collection}
\label{sec:data-collection}

In this section, we present a cost-effective data collection strategy designed to construct a multi-level, large-scale and high-quality GUI dataset, called \textbf{GUI-Lasagne}, as shown in \cref{fig:data-construction}. This dataset helps equip our models with robust abilities in GUI understanding, grounding, and navigation. The statistics of GUI-Lasagne are shown in \cref{tab:statistics-datasets}.

\subsection{Level One: Visual-Text Alignment}
\label{sec:level-1}

Visual-text alignment refers to the foundational ability of a GUI agent to perceive the visual appearance of the GUI, specifically the ability to recognize or locate text/icon elements. We collect visual-text data directly from the GUI source data. In the web scenario, we obtain website URLs from CommonCrawl~\cite{cc2024Common-Crawl} and website ranking sources. We then develop a data collection tool using Playwright\footnote{https://playwright.dev/} to gather real-world web data from the collected URLs. With this tool, we collect 755K web-page screenshots along with their DOM data, featuring diversity in both resolution and language. Additionally, we develop an icon recognition model called InternVL-Icon as an icon annotation tool, by constructing a dataset of 30K icon-caption pairs from Alibaba Icon Library\footnote{https://www.iconfont.cn/?v=20230914} and fine-tuning InternVL1.5-26B using this dataset. We then use InternVL-Icon to annotate all icons on web-pages with generated captions. As for the mobile scenario, we collect data from AitW~\cite{rawles2024AndroidintheWild}, a large-scale GUI dataset on the mobile platform. See \cref{appendix:data-collection-l1} for more collection details.

Inspired by~\citet{lee2023pix2struct} and~\citet{cheng2024seeclick}, we construct three tasks based the collected GUI source data: text2bbox, bbox2text, and bbox2dom. The \textbf{text2bbox} task prompts the model to ground the element based on the given text or icon caption. To avoid ambiguity, we provide additional contextual information for elements that appear multiple times in the screenshots. The text2bbox data is the most abundant among the three tasks, to help the model develop robust grounding capabilities. The \textbf{bbox2text} task is the inverse of the text2bbox task, training the model to recognize text and icons. The \textbf{bbox2dom} task asks the model to generate a DOM-tree corresponding to the given bounding box, as show in \cref{fig:bbox2dom}. This task helps the model learn not only to recognize text/icon elements but also to understand the GUI layout. To make sufficient use of the context length of the model, we pack multiple data pairs in each training sample for text2bbox and bbox2text task, and select the box that includes as many elements as possible for bbox2dom task. Overall, we construct a total of \textbf{1.9M} and \textbf{1.1M} training samples for the web and mobile scenarios, respectively. The data significantly enhance the GUI foundational abilities, especially the GUI grounding ability, of our SpiritSight agent. See \cref{appendix:training-data-format} for more details on sample construction.

\subsection{Level two: Visual-Function Alignment} 
\label{sec:level-2}

Visual-function alignment refers to a model's ability to locate an element based on its function. This type of data is not directly accessible from the raw GUI data. Inspired by the back-translation method~\cite{sennrich2015improving} for dataset collection in translation tasks, we leverage InternVL's image understanding capabilities to collect function grounding data. Specifically, we divide each screenshot into a 3x3 grid and describe the approximate location of the target element in text (\eg in the top-left corner of the image). Additionally, we place a bounding box around the target element in the screenshot to specify its precise location. We provide InternVL2-26B with the screenshot, the element's text content or icon caption, and the location description to prompt it to generate the element's function. Additionally, we utilize InternLM2.5-20B to enhance the quality and diversity of the generated function descriptions. These function descriptions achieve an approximate 90.9\% acceptance rate based on human judgment, which we consider sufficient for constructing the function grounding pre-training data. See \cref{appendix:data-collection-l2} for more details on data collection and validation.

Based on the strategy above, we collect \textbf{function2bbox} pairs for all interactive elements in the web scenario. In the mobile scenario, we construct the function grounding data from the GUI navigation dataset, which is be described in \cref{sec:level-3}. We use the same packing method as in the text2bbox and bbox2text tasks for efficient training, ultimately obtaining \textbf{1.5M} training samples.

\subsection{Level three: Visual GUI Navigation}
\label{sec:level-3}

We utilize the public available AitW~\cite{rawles2024AndroidintheWild} dataset to construct our GUI navigation training data. AitW is a large-scale mobile navigation dataset where each screenshot is labeled with the corresponding goal, the current step, \emph{etc.} However, as noted in AitZ~\cite{zhang2024AndroidintheZoo} and AMEX~\cite{chai2024amex}, the AitW dataset contains a number of incorrectly labeled samples. We choose to clean it with GPT-4o and employ Chain-of-Thought (CoT)~\cite{wei2022CoT} to make the judgment more accurate. Specifically, we prompt GPT-4o with the task description, the current action annotation, and screenshots from both the current and next steps. GPT-4o is then instructed to first summarize the two screenshots and identify the differences between them, then describe the current step based on these differences, and finally assess reasonability of the current action annotation. See \cref{appendix:data-collection-l3} for more details on data cleaning and quality verification.

We filter out steps identified as unreasonable by GPT-4o, resulting a final dataset of \textbf{0.64M} CoT-style GUI navigation training samples. With the collected CoT-style data, we are able to construct additional function grounding data for the mobile scenario, as mentioned in \cref{sec:level-2}, by treating the collected step descriptions as functions of the corresponding elements.

\subsection{Other Training Data}

To enhance the model's understanding of GUI content, we further collected some public datasets as a supplement, including doc/web/mobile VQA~\cite{mathew2021docvqa, chen2024guicourse, chen2021websrc, hsiao2022screenqa}, image captioning~\cite{deka2017rico, wang2021screen2words}, and mobile grounding datasets~\cite{li2020Widget-captioning, deka2017rico}, resulting in \textbf{0.59M} QA pairs for model training.

\section{Universal Block Parsing}
\label{universal-block-parsing}

\subsection{Problem Statement}

We use InternVL2~\cite{chen2024internvl2} as the base model of SpiritSight, which uses a dynamic high-resolution training approach to largely preserves the details of input images. The approach first match the optimal aspect ratio from a pre-defined set of aspect ratios to each input image, then resize and divide each image into blocks of 448×448 pixels.

\begin{figure*}[t]
  \centering
  \begin{subfigure}{0.6\textwidth}
    \centering
    \includegraphics[width=\linewidth]{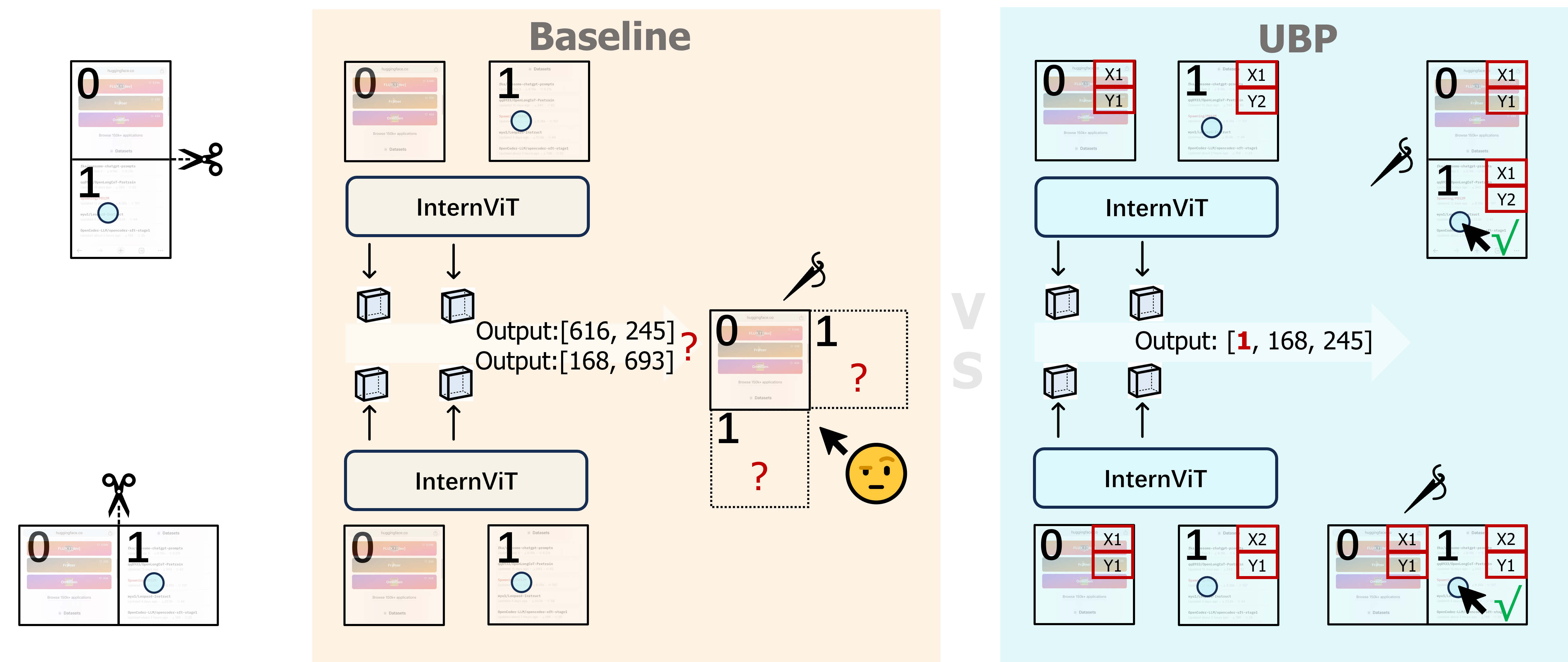}
    \caption{Baseline Block Parsing \textit{vs} Universal Block Parsing.}
    \label{fig:bbp-vs-ubp}
  \end{subfigure}
  \hfill %
  \begin{subfigure}{0.35\textwidth}
    \centering
    \includegraphics[width=\linewidth]{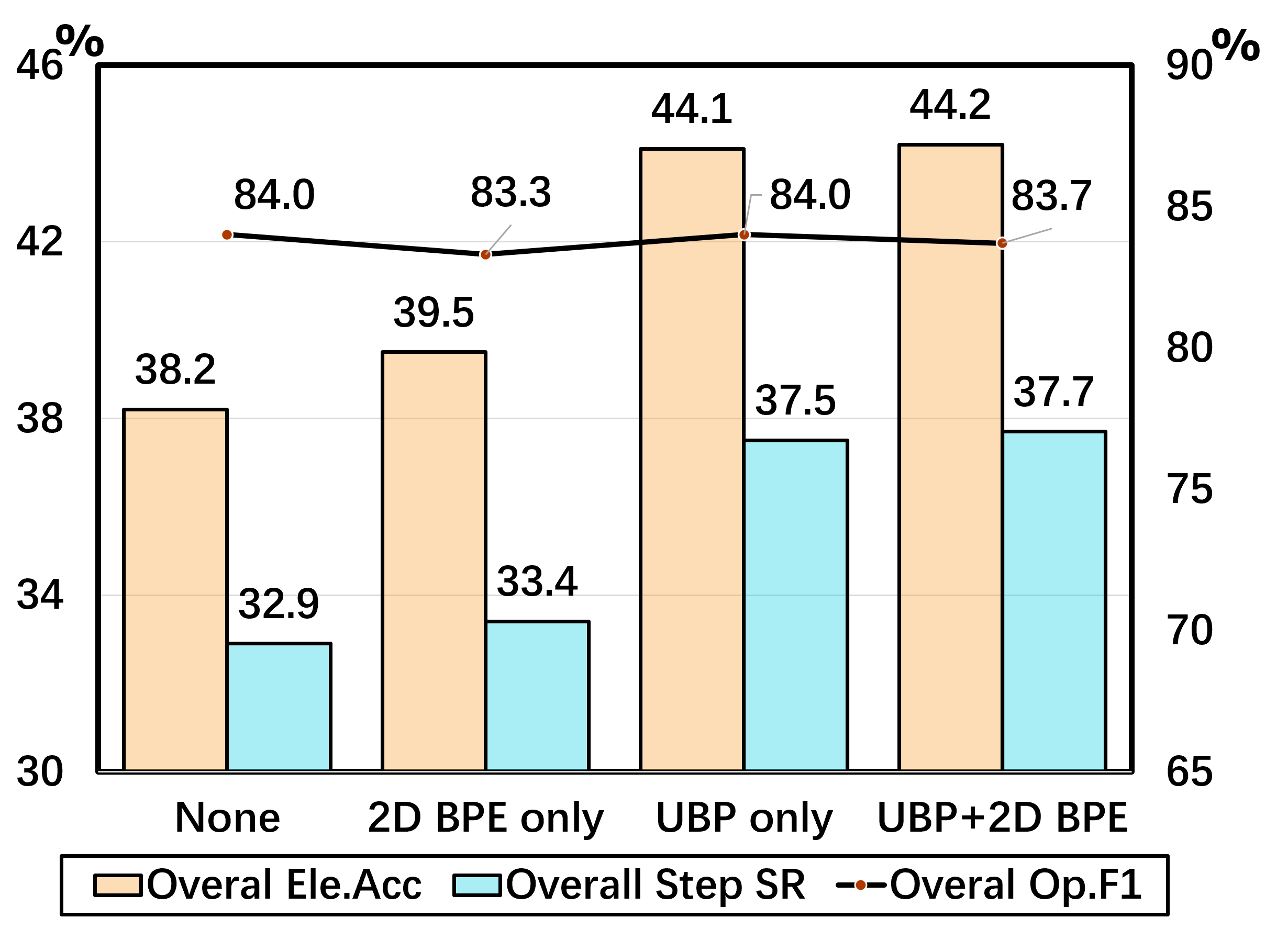}
    \caption{Performance gain with UBP.}
    \label{fig:ubp-c}
  \end{subfigure}
  \caption{(a) Comparison between baseline block parsing and our proposed UBP. (b) The results of baseline block parsing and our proposed UBP methods on Multimodal-Mind2Web benchmark. UBP improves the performance of our model. The combination of UBP and 2D-BPE achieves the best results.}
  \label{fig:ubp}
\end{figure*}

However, this approach may introduce ambiguity in grounding GUI elements. A typical example is represented in \cref{fig:bbp-vs-ubp}, with two screenshots having aspect ratios of 1:2 and 2:1, respectively. According to the dynamic high-resolution strategy, the screenshots are divided into two blocks, one vertically and the other horizontally. Suppose the target elements in each sample are located in the same relative position within the second block (block-1) after cropping. The cropped screenshots are then fed into InternVL and the model is expected to predict different locations from these two identical inputs. We refer to this situation as the locational ambiguity problem.

Generally, a point $\mathbf{p}$ is expressed in the global coordinate system as
\begin{equation}
    \mathbf{p}=[x, y]
\end{equation}
where $x$ and $y$ represent the horizontal and vertical coordinate values of the point in the original image, respectively. Assume that the image is resized and divided into $n_w \times n_h$ blocks. We can express the point $\mathbf{p}$ in the corresponding block as $\mathbf{p'}$:
\begin{align}
\small
    \begin{cases}
        b_x = \lfloor \frac{x}{w_{block}} \rfloor \\
        b_y = \lfloor \frac{y}{h_{block}} \rfloor
    \end{cases}
    \begin{cases}
        x' = x \bmod w_{block} \\
        y' = y \bmod h_{block}
    \end{cases}
\end{align}
\begin{equation}
    \mathbf{p'}=[b_x, b_y, x', y']
\end{equation}
where $w_{block}$ and $h_{block}$ (both set to 448 in our experiments) represent the width and height of each block, respectively. $b_x$ and $b_y$ represent the horizontal and vertical indices of the block containing $\mathbf{p'}$, and $x'$ and $y'$ represent the coordinates of $\mathbf{p'}$ within this block. The blocks are flattened into a sequence before being fed into the model, thus we have
\begin{equation}
\label{eq:UBP}
    \mathbf{p''} = [b_i, x', y']
\end{equation}
where $b_i$ represents the index of the block within the sequence, satisfying:
\begin{equation}
\label{eq:flatten}
    b_{i} = b_y \cdot n_w + b_x
\end{equation}
The model is trained to approximate a mapping $f: \mathbf{p''} \rightarrow \mathbf{p}$, which is inherently a multi-valued function. For example, when the input $\mathbf{p''}$ is $[1, 168, 245]$, possible values for $p$ include $[168, 693]$ when $n_w=1$, or $[616, 245]$ when $n_w=2$, as shown in \cref{fig:bbp-vs-ubp}.

\subsection{Method}

The ambiguity primarily arises from the flattening operation, as shown in \cref{eq:flatten}, which results in the loss of the spatial relationship between blocks. One solution is to input an additional thumbnail, but this may lead to extra computational and memory overhead. We propose to solve this positional ambiguity with two steps. Firstly, we introduce a 2D Block-wise Position Embedding (2D-BPE)~\cite{ye2023ureader} by adding two position embeddings to each block to capture spatial information. Secondly, we introduce a \textbf{Universal Block Parsing (UBP)} method, which replaces the global coordinate representation with a block-specific coordinate representation. Specifically, we express a point as \cref{eq:UBP}. In this case, the model is trained to approximate a injective mapping $f: \mathbf{p''} \rightarrow \mathbf{p''}$, thereby resolving the ambiguity problem. During the model inference, the global coordinate of this point can be computed in post-processing as follows:
\begin{align}
\small
\begin{cases}
x = x' + (b_i \mod n_w) \cdot w_{block} \\
y = y' + \lfloor \frac{b_i}{n_w} \rfloor \cdot h_{block}
\end{cases}
\end{align}

Overall, our UBP method ensures a clear mapping of positional information between the model's inputs and outputs, thereby improving the model's grounding capability.

\section{Settings}

\subsection{Implementation Details}
\label{implementation-details}
We use InternVL2 (2B, 8B and 26B)~\cite{chen2024internvl2} as the base models and train them with two stages: continual pre-training and fine-tuning. During the pre-training stage, we use all the GUI-Lasagne dataset mentioned in \cref{sec:data-collection}. Different prompts are designed for different training tasks to avoid task confusion. See \cref{appendix:training-data-format} for the prompts. We unfreeze the vision encoder, decoder, and MLP layer of InternVL. The learning rate is set to 1e-4/1e-4/5e-5 for InternVL-2B/8B/26B, respectively, and the batch size is set to 1024. We get \textbf{SpiritSight-Base} model after pre-training and then fine-tune it on multiple downstream tasks individually. More details are shown in \cref{appendix:implementation-benchmark} and \cref{appendix:implementation-ablation}.

\subsection{Benchmarks \& Metrics}

To assess SpiritSight’s capability in diverse real-world environments, we evaluate SpiritSight on six benchmarks covering various GUI platforms and tasks. Multimodal-Mind2Web~\cite{deng2024mind2web}, ScreenSpot~\cite{cheng2024seeclick}, AMEX~\cite{chai2024amex}, GUIAct~\cite{chen2024guicourse}, AndroidControl~\cite{li2024androidcontrol}, and GUI-Odyssey~\cite{lu2024GUI-Odyssey}. For Multimodal-Mind2Web and ScreenSpot, we use the same data pre-processing and metrics as SeeClick~\cite{cheng2024seeclick} uses. For GUIAct, we evaluate SpiritSight on the web-single and web-multi sub-sets and report step success rate (Step SR). For AndroidControl, we evaluate SpiritSight on the High-Level (HL) and Low-Level (LL) tasks and report step accuracy. For AMEX and GUI-Odyssey, we report the action matching score (AMS) defined in AitW~\cite{rawles2024AndroidintheWild}. These metrics are similar in that they all measure the single-step accuracy. Refer to \cref{appendix:metrics} for more information about the metrics.

\section{Experiment}
\subsection{Advanced Vision-based GUI Agent}
In this section, we compare the performance of SpiritSight with other advanced methods across various input modalities and test configurations on Multimodal-Mind2Web~\cite{deng2024mind2web}, a classic and high-quality benchmark for GUI navigation. The results are shown in \cref{sota-mind2web}. Methods that use top-50 candidates as input perform the best. This is evident, as the assistance of candidate elements can significantly reduce the decision space. However, such methods are not particularly feasible in practice.

The results indicate that SpiritSight significantly outperforms all methods that do not rely on any candidate element, including vision-based methods, language-based methods, and even vision-language-based methods. This demonstrates strong capabilities of SpiritSight in Web GUI navigation tasks. It should be noted that SpiritSight achieves a significant advantage in the Ele.Acc metric compared to other vision-based methods, which can be attributed to the specially constructed visual grounding training data and the proposed UBP approach.

\subsection{Strong Cross-Platform Compatibility}
We evaluated SpiritSight on other GUI navigation benchmarks across various platforms and compare it with state-of-the-art (SOTA) GUI agents as shown in \cref{sota-newer-benchmarks}. SpiritSight demonstrated the best performance on these benchmarks, showing its strong capabilities across various platforms. OdysseyAgent uses additional historical screenshot images as input, yet SpiritSight achieves comparable results.

We also evaluated SpiritSight on ScreenSpot, a function grounding benchmark. As shown in \cref{sota-screenspot-small}, SpiritSight performs well across all three platforms, showcasing its cross-platform capability and robust grounding ability. Notably, SpiritSight models are trained specifically for GUI navigation tasks, with no training data intentionally aligning with ScreenSpot is collected. There remains room for further improvement in SpiritSight’s performance on the ScreenSpot benchmark. We further assess the foundational ability of SpiritSight, the visual grounding ability, on our custom-constructed testing dataset. See \cref{appendix:gui-grounding-abilities} for more details.

\begin{table}[t]
\renewcommand{\arraystretch}{1.3}
\begin{center}
\small
\setlength{\tabcolsep}{2pt} %
\begin{tabular}{ccccccc}
\hline
\multirow{2}{*}{Agent} &
  Odyssey &
  AMEX &
  \multicolumn{2}{c}{AndroidCtrl} &
  \multicolumn{2}{c}{GUIAct} \\ \cline{2-7} 
               & High & High & High & Low  & Multi & Single \\ \hline\hline
GPT-4o\cite{gpt4o}          & 20.4\% & - & 21.2\% & 28.4\% & -  & 41.8\% \\
Previous SOTA  & 74.3\% & 70.7\% & 64.8\% & 80.0\% & 45.4\% & 74.9\% \\ \hline
SpiritSight-2B & 72.3\% & 74.5\% & 64.9\% & 86.3\% & 45.5\% & 76.0\% \\
SpiritSight-8B & \textbf{75.8}\% & \textbf{80.7}\% & \textbf{68.1\%} & \textbf{87.6\%} & \textbf{49.3\%} & \textbf{78.2\%} \\ \hline
\end{tabular}
\end{center}
\caption{Results of SpiritSight on four recently published GUI navigation benchmarks. 'High' and 'Multi' indicates high-level tasks, 'Low' and 'Single' indicates single-stepped tasks. The SOTA results of GUI-Odyssey, AMEX, AndroidControl and GUIAct come from OdysseyAgent~\cite{lu2024GUI-Odyssey}, SPHINX-GUI-Agent~\cite{chai2024amex}, fine-tuned PaLM-2S~\cite{li2024androidcontrol} and MiniCPM-GUI~\cite{chen2024guicourse}, respectively.}
\label{sota-newer-benchmarks}
\renewcommand{\arraystretch}{1.15}
\end{table}

\subsection{Recognition and Grounding as Priors for GUI Navigation}

To verify the significance of the three levels of GUI-Lasagne data, we progressively removed level-3, level-2, and level-1 data from the training set during the pre-training stage and evaluate SpiritSight-8B on Multimodal-Mind2Web. The results are shown by the blue line in \cref{fig:data-ablation-a}. It can be seen that each level of data contributes to improving Step SR. While the tasks in level-1 data differ the most from web navigation task compared to the other two levels, they provide an effective initialization for the pre-trained model. Although the level-3 data is constructed from the mobile scenario, it also assists in web-based GUI navigation tasks. This indicates that the joint learning strategy helps SpiritSight develop strong navigation abilities across different GUI environments with limited resources.

We also conducted ablation experiments to evaluate the effectiveness of data cleaning and CoT construction strategies on the level-3 data, as shown by the orange line in \cref{fig:data-ablation-a}. It can be observed that training in a CoT manner effectively improve the model's GUI navigation capabilities, while the impact of data cleaning strategy is relatively small. This may be due to the fact that erroneous samples are relatively tolerated during the pre-training phase of large models. The difference in results of "w/ CoT" and "w/ Level 1+2+3" is due to the different implementation for fine-tuning stage. See \cref{appendix:implementation-ablation} for more details.

To evaluate the effectiveness of our GUI-Lasagne dataset, we train InternVL-2 (8B) on SeeClick training data~\cite{cheng2024seeclick} using full parameter training and LoRA training, resulting in SeeClick(InternVL-Full) and SeeClick(InternVL-LoRA), respectively. We then fine-tune and evaluate the two models on the Multimodal-Mind2Web benchmark. The results are shown by the dashed line in \cref{fig:data-ablation-a}, where we also present the result of the original SeeClick model. SeeClick(InternVL-LoRA) performs better than the original SeeClick model, which may be attributed to the dynamic high-resolution approach used by InternVL that preserves the details of input images. SeeClick(InternVL-Full) performs the worst, as the scale of the SeeClick training data is not large enough for the model to converge during full parameter training. SpiritSight agent that trained on only GUI-Lasagne level-1 data outperforms all three models trained on SeeClick data, demonstrating the effectiveness of GUI-Lasagne dataset.

\begin{table}[t]
\renewcommand{\arraystretch}{1.3}
\begin{center}
\small
\begin{tabular}{ccccc}
\hline
\multirow{2}{*}{Agent} & \multirow{2}{*}{\begin{tabular}[c]{@{}c@{}}Model\\ Size\end{tabular}} & \multicolumn{3}{c}{ScreenSpot} \\ \cline{3-5} 
               &      & Web  & Mobile & Desktop \\ \hline\hline
GPT4V\cite{achiam2023gpt4}          & -    & 5.0\%  & 7.5\%    & 4.6\%     \\
Qwen-VL\cite{bai2023qwen-vl}        & 9.6B & 3.0\%  & 7.2\%    & 5.4\%     \\ \hline
Fuyu\cite{fuyu-8b}           & 8B   & 19.2\% & 21.2\%   & 18.3\%    \\
CogAgent\cite{hong2024cogagent}       & 18B  & 49.5\% & 45.5\%   & 47.1\%    \\
SeeClick\cite{cheng2024seeclick}       & 9.6B & 44.1\% & 65.0\%   & 51.1\%    \\ \hline
SpiritSight-2B & 2B   & 63.6\% & 62.5\%   & 61.8\%    \\
SpiritSight-8B & 8B   & \textbf{68.3\%} & \textbf{68.4\%}   & \textbf{62.9\%}    \\ \hline
\end{tabular}
\end{center}
\caption{Results of SpiritSight and other vision-based methods on ScreenSpot Benchmark.}
\label{sota-screenspot-small}
\renewcommand{\arraystretch}{1.15}
\end{table}

\subsection{Better Grounding Ability from UBP}
To verify the effectiveness of UBP on grounding task, we use LoRA for resource efficiency to fine-tune InternVL in 4 different settings (original as baseline, 2D-BPE, UBP, and 2D-BPE\&UBP), respectively. We then evaluate these models on Multimodal-Mind2Web benchmark, as shown in \cref{fig:ubp-c}. It can be seen that UBP shows a significant advantage in Ele.Acc compared to baseline, while the results of Op.F1 show little variation across the 4 settings. This indicates that UBP improves the performance of GUI agent primarily by enhancing the grounding ability. Finally, the combination of UBP and 2D-BPE achieves the best results. This indicates that UBP is compatible with 2D-BPE, leading to better performance.

\subsection{Scaling Effects on Dataset and Model Size}
\begin{figure*}[t]
  \centering
  \begin{subfigure}{0.3\textwidth}
    \centering
    \raisebox{-0.03\height}{\includegraphics[width=\linewidth]{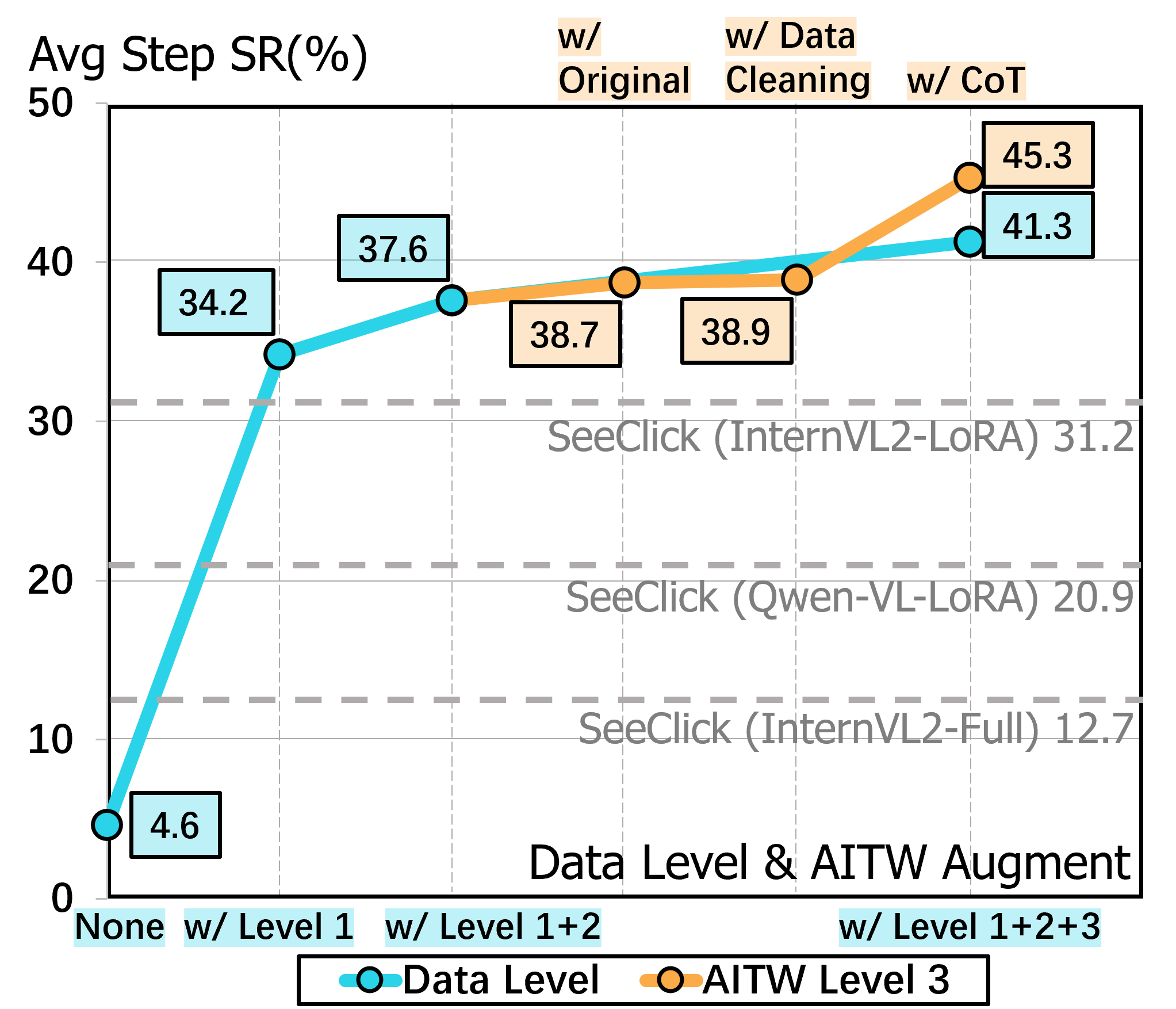}}
    \caption{Effect of three data levels and data augmentation for level-3 data}
    \label{fig:data-ablation-a}
  \end{subfigure}
  \hfill %
  \begin{subfigure}{0.3\textwidth}
    \centering
    {\includegraphics[width=\linewidth]{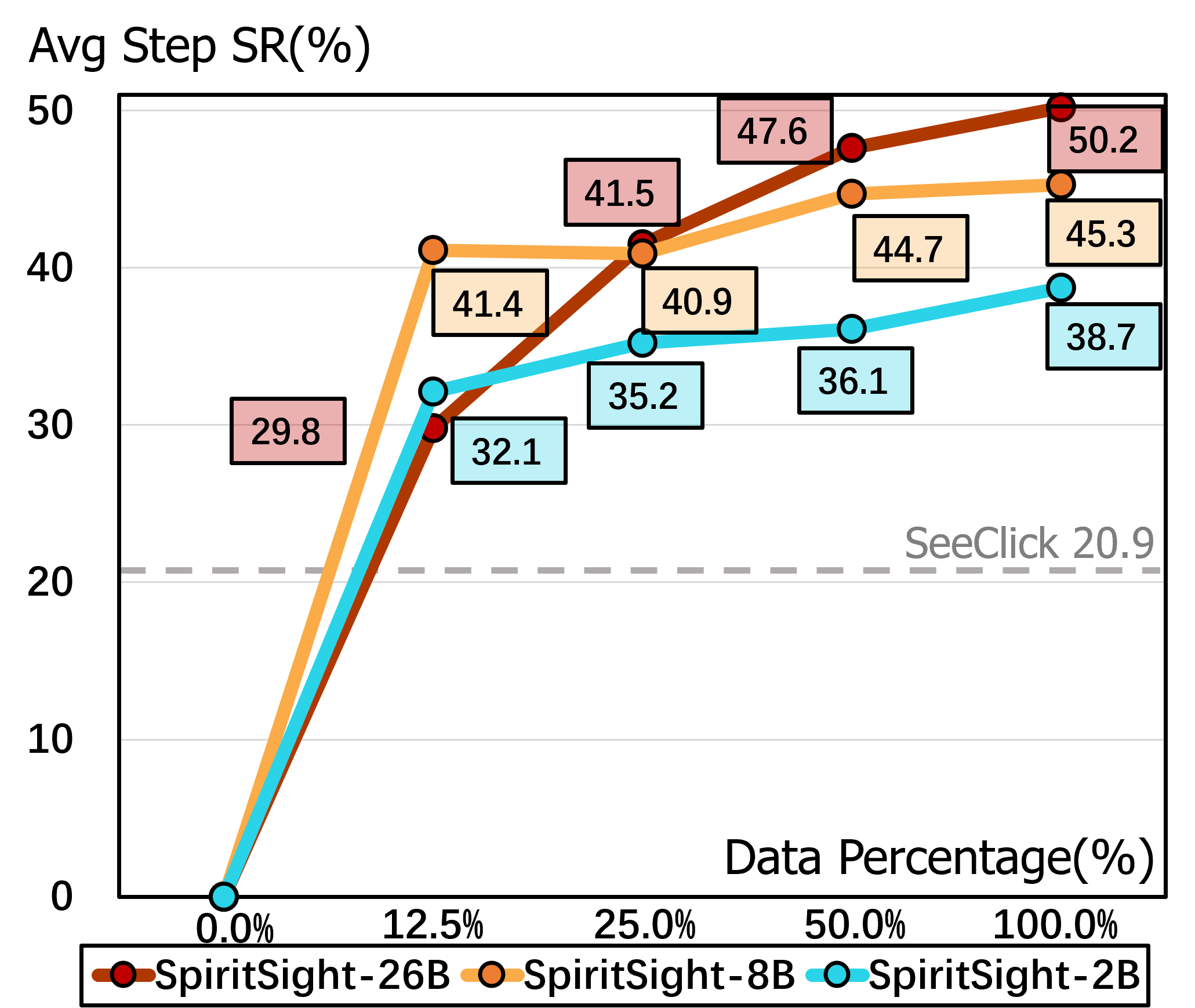}}
    \caption{Ablation Study: Effect of training data percentages on model performance.}
    \label{fig:data-ablation-b}
  \end{subfigure}%
  \hfill %
  \begin{subfigure}{0.3\textwidth}
    \centering
    {\includegraphics[width=\linewidth]{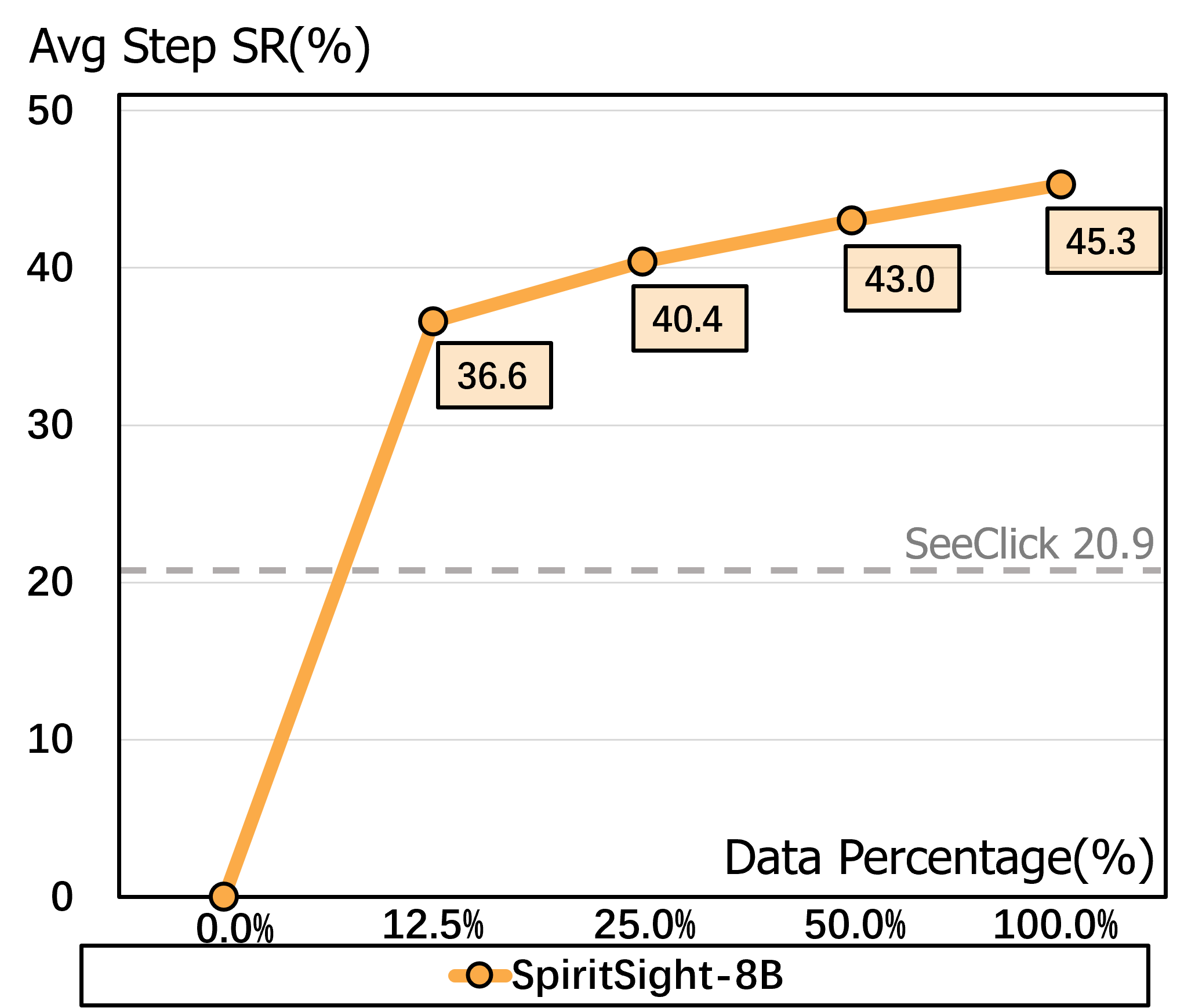}}
    \caption{Ablation Study: Effect of data percentages on downstream task.}
    \label{fig:data-ablation-c}
  \end{subfigure}%
  \caption{Ablation studies on GUILasagne dataset and scaling effects in the Multimodal-Mind2Web benchmark.}

  \label{fig:data-ablation}
\end{figure*}

We explore the impact of pre-training dataset and model size on SpiritSight using Multimodal-Mind2Web benchmark. The results are shown in \cref{fig:data-ablation-b}. SpiritSight-2B outperforms SeeClick~\citep{cheng2024seeclick} by using just 1/8 of the pre-training dataset. This impressive performance comes from the high quality and grounding-focus nature of GUI-Lasagne data. The performance improves as the size of dataset increases, demonstrating the significance of collecting large-scale data. SpiritSight-2B reaches saturation with a smaller amount of pre-training data, while SpiritSight-26B appears to have further potential for improvement, which aligns with the scaling law of LLMs and VLMs.

We also evaluate the ability of SpiritSight to transferring to downstream GUI agent tasks. We fine-tune SpiritSight-Base (8B) on various proportions of the Multimodal-Mind2Web training data and show the results in \cref{fig:data-ablation-c}. It is noticed that SpiritSight achieves 36.6\% Step SR with only 1/8 of the fine-tuning data, showing strong transferability to GUI navigation tasks.

\subsection{Effective Transfer to other languages}
Exploring the cross-lingual capabilities of GUI agents is highly beneficial for their application in non-English environments. We split the training and testing sets of GUIAct (web-multi) dataset into English and Chinese partitions, respectively. We then fine-tune SpiritSight-Base (8B) on two sets of data: the entire training set (English\&Chinese) and the English-only training set. The results are shown in \cref{ablation-zeroshot-multi-language}.

Under the English\&Chinese configuration, SpiritSight achieves comparable results on both the English and Chinese testing sets, despite having fewer Chinese samples in the pre-training dataset than English ones. Under the English-only configuration, SpiritSight achieves 24.5\% Step SR on the Chinese testing set, reaching half of the English\&Chinese performance. The zero-shot capability of SpiritSight in Chinese results from the small but effective foundational Chinese data included in the pre-training stage.

This experiment offers a paradigm for applying GUI agents to non-English environments: by collecting (1) free web and mobile GUI information from the target language environment (level 1 \& level 2 data), and (2) a small amount of high-quality GUI navigation data at low cost (level 3 data). With this language transferring strategy, the same capabilities as in the English environment can be achieved in non-English environments with minimal extra costs.

\begin{table}[t]
\renewcommand{\arraystretch}{1.3}
\begin{center}
\small
\begin{tabular}{cccc}
\hline
SFT Data & Overall & Chinese & English \\ \hline\hline
{\color[HTML]{333333} English+Chinese} &
  49.3\% &
  49.3\% &
  49.2\% \\
{\color[HTML]{333333} English} &
  35.0\% &
  24.5\% &
  48.6\% \\ \hline
\end{tabular}
\end{center}
\caption{Results of SpiritSight on GUIAct (web-multi), trained with different language datasets.}
\label{ablation-zeroshot-multi-language}
\renewcommand{\arraystretch}{1.15}
\end{table}

\section{Limitation}

As SpiritSight is a vision-based GUI agent, it constantly requires access to screenshots which may contain personal information or other sensitive data. Users and system providers should carefully manage the system privileges granted to SpiritSight agent to mitigate potential privacy and security risks.

\section{Conclusion}
In this paper, we propose an advanced vision-based end-to-end GUI agent, SpiritSight, with high generalization across multiple GUI platforms. We construct an efficient multi-level, large-scale, high-quality GUI pre-training dataset to equip SpiritSight with robust GUI perception, grounding and understanding capabilities. Additionally, we introduce a UBP method to resolve ambiguity in dynamic high-resolution inputs during model training, further enhancing the ability of SpiritSight to ground GUI objects. As a result, SpiritSight achieves strong performance in numerous GUI navigation benchmarks, demonstrating significant potential for practical deployment in real-world applications.

{\small
\bibliographystyle{ieeenat_fullname}
\bibliography{main}
}

\ifarxiv \clearpage \appendix
\section{Extended Related Work}
\label{appendix:related-work}

\subsection{Large-scale Language Models}
In recent years, large language models (LLMs)~\cite{radford2018gpt1, devlin2018bert, raffel2020T5, xu2021commonsenseqa, wu2023bloomberggpt, nijkamp2022codegen, roziere2023codellama, yu2023metamath, wang2023mathcoder, ying2024internlm-math, shao2024deepseekmath, wei2022CoT, wei2022emergent, pan2023logic-lm} have demonstrated remarkable capabilities in the field of Natural Language Processing (NLP), encompassing natural language generation, commonsense knowledge question-answering, code completion, mathematical computation, and logical reasoning. LLM have also demonstrated strong decision-making capabilities, laying the foundation for the emergence of GUI agents.

\subsection{Visual Large-scale Language Models}
With the development of large language models, numerous works~\cite{bai2023qwen-vl, wang2023cogvlm, lin2023sphinx, li2024llava-next, chen2024internvl2, zhang2024internlmxcomposer25, yao2024minicpm, jin2309unified} have proposed vision language models (VLMs) to bring the capabilities of language models into the visual domain. CLIP~\cite{hafner2021clip} uses contrastive learning to align vision and language features, while BLIP~\cite{li2022blip} and BLIP-2~\cite{li2023blip2}build on this by adding a language decoder, enabling the models to perform image-grounded text generation. InternVL~\cite{chen2024internvl} attempts to scale the parameters of vision encoder up to 6 billion, significantly enhancing the model's ability to perceive visual input. LLaVA~\cite{liu2023llava} and Sphinx~\cite{lin2023sphinx} improve the models' understanding and chat abilities through instruction tuning and multitask learning, respectively. Beyond general domains, OCR-Free~\cite{kim2022ocr-free} methods use an encoder-decoder architecture to achieve end-to-end visual document understanding. This demonstrates the significant potential of VLMs in GUI navigation tasks.

\subsection{GUI Agent Benchmarks}
GUI agents have seen rapid development in recent years, with many types of benchmarks emerging. MiniWoB~\cite{shi2017MiniWoB}, MiniWoB++~\cite{liu2018MiniWoB++}, and WebShop~\cite{yao2022webshop} are early classic GUI navigation benchmarks. However, the data in these benchmarks is synthetically generated, which creates a slight gap compared to real-world data. AitW~\cite{rawles2024AndroidintheWild} is a large-scale real-world dataset designed for mobile navigation tasks, and Mind2Web~\cite{deng2024mind2web} is a high-quality benchmark for web navigation that provide evaluation across three scenario: cross-task, cross-website, and cross-domain. ScreenSpot~\cite{cheng2024seeclick} is a functional grounding benchmark that covers mobile, web, and desktop scenarios. GUIAct~\cite{chen2024guicourse}, AMEX~\cite{chai2024amex}, AndroidControl~\cite{li2024androidcontrol}, and GUI Odyssey~\cite{lu2024GUI-Odyssey} are newly released benchmarks designed for web and mobile environments, respectively. They are highly reliable benchmarks as they are annotated by humans and have undergone further validation. In this paper, we evaluate SpiritSight on six benchmarks across different GUI platforms and tasks: Multimodal-Mind2Web~\cite{deng2024mind2web}, ScreenSpot~\cite{cheng2024seeclick}, GUIAct~\cite{chai2024amex}, AMEX~\cite{chai2024amex}, AndroidControl~\cite{li2024androidcontrol}, and GUI Odyssey~\cite{lu2024GUI-Odyssey}. Overview of these benchmarks is shown in \cref{tab:statistics-benchmarks}

\section{Task Formulation}
\label{appendix:task-formulation}

For a given GUI platform, we first obtain an action space $\mathcal{A}$ that contains all operable actions. Given the task description $\mathcal{T}$, the previous actions $\mathcal{H}=\{a_1, a_2, ..., a_{t-1}\}$, the action space $\mathcal{A}$ and the current screenshot $o_t$, the agent is expected to infer the optimal action $a^*_t$ that maximizes the expected future reward. The inference process is guided by a policy $\pi$, as shown below, which maps the current context to a probability distribution over the action space $\mathcal{A}$. Here, $a$ denotes a specific action selected from the action space $\mathcal{A}$.
\begin{equation}
a^*_t \sim \pi(a|\mathcal{T}, \mathcal{H}, \mathcal{A}, o_t), \quad a \in \mathcal{A}
\end{equation}

We propose a hierarchical decomposition of the policy to manage the complexity of action inference. Initially, we define $s$ as the natural language description (\eg Click on the login button.) of action $a$ (\eg CLICK(132, 243)). We decompose the overall policy $\pi$ into a step inference policy $\pi_s$ and an action inference policy $\pi_a$ as \cref{formula:decompose1}. The step inference policy $\pi_s$ selects the optimal $s$ based on the current context. Once $s$ is determined, the action inference policy $\pi_a$ predicts the corresponding action $a$ from the action space $\mathcal{A}$.
\begin{equation}
\label{formula:decompose1}
\pi(a|\mathcal{T}, \mathcal{H}, \mathcal{A}, o_t) = \pi_s(s|\mathcal{T}, \mathcal{H}, o_t) \cdot \pi_{a}(a|s, \mathcal{A})
\end{equation}

Further, we decompose $\pi_a$ into $\pi_{pos}$ and $\pi_{attr}$ as \cref{formula:decompose2}. Here, $a_{pos}$ denotes to the positional aspect of the action, typically the coordinates where the action is performed (\eg (132, 243)), while $a_{attr}$ denotes the non-positional aspects, such as the action type (\eg click) or additional parameters like specific input text.
\begin{equation}
\label{formula:decompose2}
\pi_{a}(a|s, \mathcal{A}) = \pi_{pos}(a_{pos}|s, \mathcal{A}) \cdot \pi_{attr}(a_{attr}|s, \mathcal{A})
\end{equation}

Based on \cref{formula:decompose1} and \cref{formula:decompose2} we have
\begin{equation}
\begin{aligned}
\pi(a|\mathcal{T}, \mathcal{H}, \mathcal{A}, o_t) = 
&\; \pi_s(s|\mathcal{T}, \mathcal{H}, o_t) \cdot \\
&\; \pi_{pos}(a_{pos}|s, \mathcal{A}) \cdot \\
&\; \pi_{attr}(a_{attr}|s, \mathcal{A})
\end{aligned}
\end{equation}
It is easy for vision-based agents to learn the step inference policy $\pi_s$, as recent VLMs excel at image understanding and reasoning. Learning the non-positional inference policy $\pi_{attr}$ is also manageable, since the non-positional aspects of an action can be directly inferred from the natural language step. For example, an action like "\textit{INPUT('Copenhagen')}" can be directly inferred from a step such as "\textit{Input 'Copenhagen' into the arrival input box}". The primary challenge lies in learning the positional sub-policy $\pi_{pos}$ as discussed in \cref{sec:related_works}. To address this challenge, we construct a large scale dataset focused primarily on grounding tasks to facilitate learning accurate positional actions.

\begin{figure*}[t]
\centering
\includegraphics[width=0.8\textwidth]{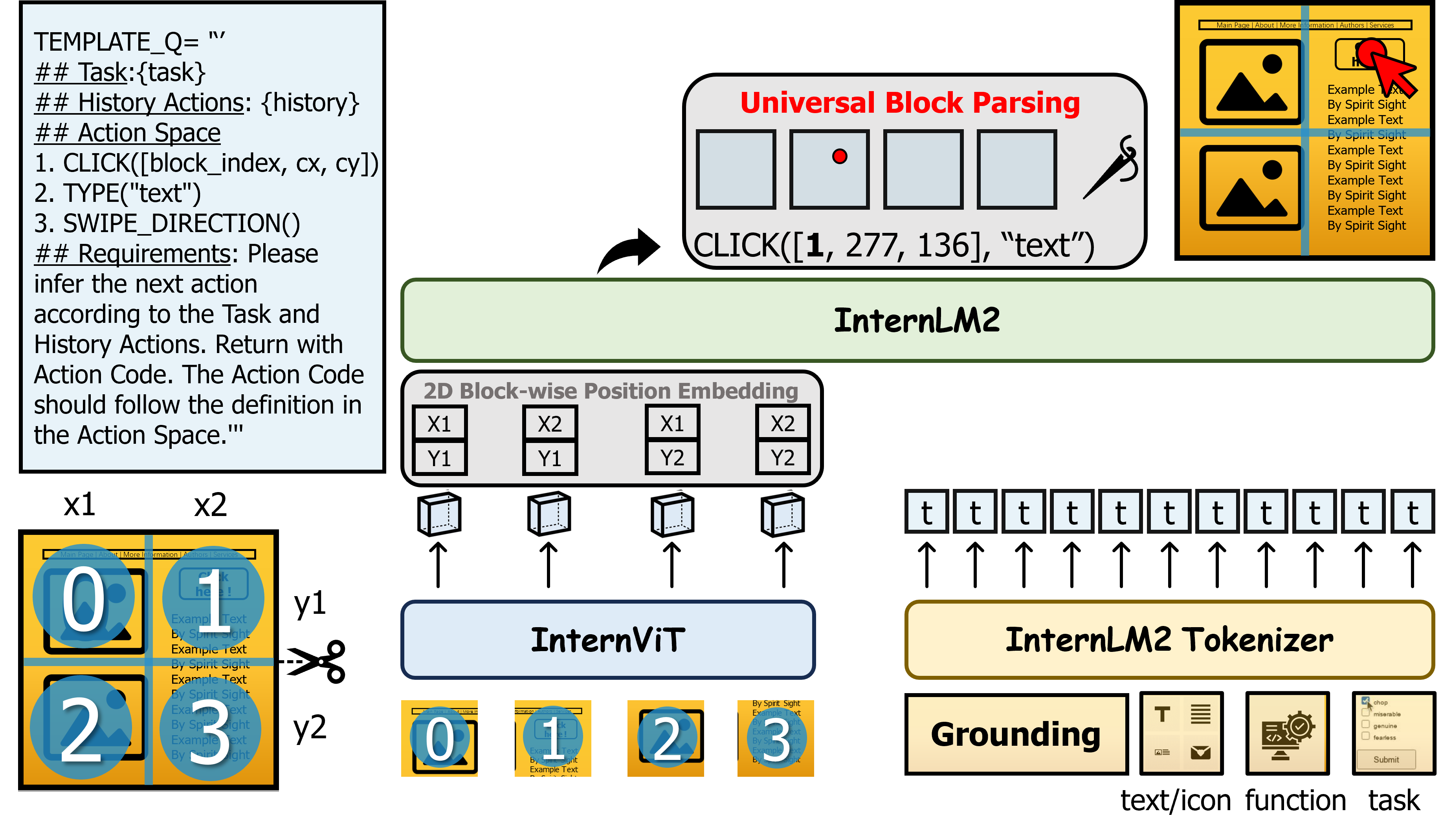}
\caption{The overall architecture of SpiritSight. SpiritSight is pre-trained on GUI-Lasagne, a large-scale, multi-level, high quality GUI dataset. The UBP method solves the ambiguity in Dynamic High-Resolution input during model training.}
\label{fig:overall-architecture}
\end{figure*}

\section{Overall Architecture}
\label{sec:overall_architecture}

We build our model based on the pre-trained InternVL2, a family of advanced and open-sourced VLMs. We chose InternVL for the following reasons: (1) The large-scale and high-performance vision encoder is more capable to handle the text-rich GUI environment. (2) The dynamic resolution strategy largely preserves the details of the input screenshots, allowing for enhanced perception of fine-grained text and icon information. We take the advantage of large-scale InternViT with a large-scale GUI dataset described in \cref{sec:data-collection}. We further propose a Universal Block Parsing (UBP) method to resolve the ambiguity problem brought by dynamic resolution in \cref{universal-block-parsing}. 

The architecture of SpiritSight is depicted in \cref{fig:overall-architecture}. To begin with, the input image is the GUI screenshot. According to the dynamic resolution algorithm of InternVL, an appropriate ratio of input image is decided. Then, the image is divided into several blocks, each with a unique index, in preparation for the post-processing phase of our UBP method. These image blocks will be flattened into sequences before being sent to the vision encoder, which results in the loss of their 2D spatial relation. To address this problem, we introduce 2D Block-wise Position Embedding (2D-BPE) method, which maintains the blocks' 2D spatial relation by adding a row embedding and column embedding to each block. Afterwards, the embedded image features, along with the task objective, the action space and the history actions are processed by the InternLM2 decoder to infer the action code. Finally, the action and corresponding coordinate for operation is obtained by the UBP parser.

\begin{figure*}[t]
  \centering
    \includegraphics[width=0.9\linewidth]{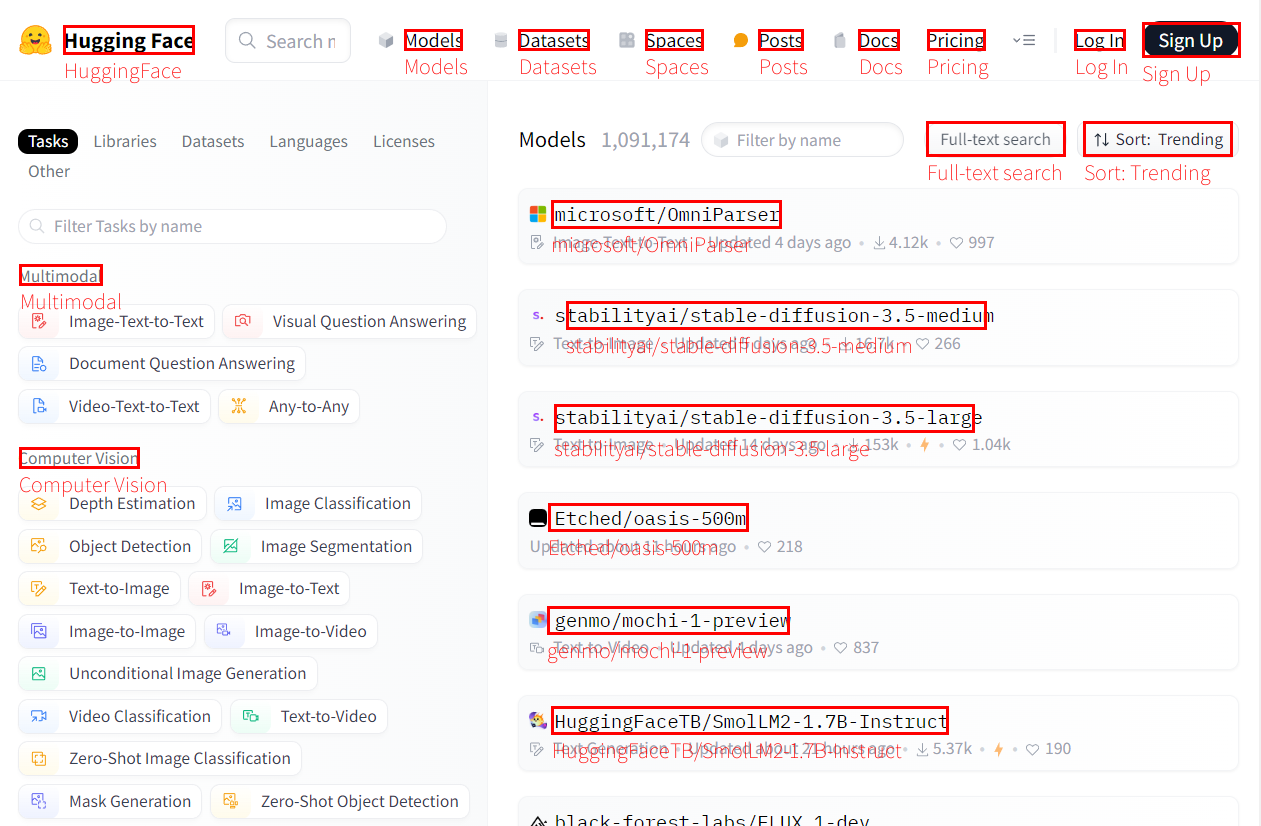}
    \caption{Visualization results of SpiritSight-2B on our custom text2bbox test set. The red boxes represent the generated results and the text next to it represent the text prompt.}
  \label{fig:vis-text2bbox}
\end{figure*}

\section{Experiments}

\subsection{Implementation Details for Benchmark Experiments}
\label{appendix:implementation-benchmark}
We use InternVL2 (2B, 8B and 26B)~\cite{kim2022ocr-free, chen2024internvl2} as the base models and train them with two stages: continual pre-training and fine-tuning. 

\textbf{Pre-training Stage.} We train all the GUI-Lasagne datasets mentioned in the \cref{sec:data-collection} simultaneously. Different prompts are designed for different training tasks to avoid task confusion. See \cref{appendix:training-data-format} for the detailed prompt content. We unfreeze the vision encoder, decoder, and MLP layer of InternVL. The learning rate is set to 1e-4/1e-4/5e-5 for InternVL-2B/8B/26B, respectively, and the batch size is set to 1024. We get \textbf{SpiritSight-Base} models after pre-training.

\textbf{Fine-Tuning Stage.} We fine-tune SpiritSight-Base models in several downstream tasks separately. We define a distinct action space for each task to prevent action confusion. The max number of history actions is set to 5 to prevent excessive overload. For ScreenSpot benchmark, we follow the data proportions from~\citet{cheng2024seeclick}, using part of the level-1 and level-2 data of GUI-Lasagne, and the data from~\citet{li2020Widget-captioning, deka2017rico, wang2021screen2words} to train the entire model. For other GUI navigation benchmarks, we first train the entire model for 1 epoch using the level-3 data of GUI-Lasagne and the training data corresponding to each benchmark, then fine-tune the model for 1 epoch on the benchmark-specific training data using LoRA~\cite{hu2021lora}. While training the entire model, the learning rate is set to the same as pre-training, and the batch size is 1024. During fine-tuning, the learning rate is set to 5e-5, the batch size is 64, with the alpha of vision encoder and LLM decoder set to 32 and 64, respectively.

\subsection{Implementation Details for Ablation Study}
\label{appendix:implementation-ablation}

\textbf{Recognition and Grounding as Priors for GUI Navigation.}
To verify the significance of the three levels of GUI-Lasagne data, we progressively removed level-3, level-2, and level-1 data from the training set during the pre-training stage and evaluate models on Multimodal-Mind2Web benchmark. During the fine-tuning stage, we train the SpiritSight-Base model only on the Multimodal-Mind2Web training data using LoRA, without training the whole models on level-3 data of GUI-Lasagne, as level-3 data is excluded from the pre-training datasets. The results are shown by the blue line in \cref{fig:data-ablation-a}.

We also conducted ablation experiments to evaluate the effectiveness of data cleaning and CoT construction strategies on the level-3 data, as shown by the orange line in \cref{fig:data-ablation-a}. We use the same setting as in the benchmark experiments, except with different versions of level-3 data (original version, data cleaning version and CoT version).

\textbf{Better Grounding Ability from UBP.}
To verify the effectiveness of UBP on grounding task, we use LoRA for resource efficiency to fine-tune InternVL in 4 different settings (original as baseline, 2D-BPE, UBP, and 2D-BPE\&UBP), respectively. First, We fine-tune InternVL1.5 (26B) on 10\% of our GUI-Lasagne dataset. The alpha is set to 64. Then, we fine-tune the model on the Multimodal-Mind2Web training data. The alpha is set to 16 and 32 for vision encoder and LLM decoder, respectively.

\textbf{Scaling Effects on Dataset and Model Size.}
We explore the impact of pre-training dataset and model size on SpiritSight using Multimodal-Mind2Web benchmark. The training strategies are same as in the benchmark experiments, except with different scale of pre-training data.

\textbf{Effective Transfer to other languages.}
We split the training and test sets of GUIAct(web-multi) dataset into English and Chinese partitions, respectively. We fine-tune SpiritSight-Base (8B) on two sets of data: the entire training set (English\&Chinese) and the English-only training set. Other training strategies are same as in the benchmark experiments.

\begin{table*}[t]
\begin{center}
\small
\begin{tabular}{cccccc}
\hline
Benchmarks          & Platforms       & Task                 & Metric                  & \# Test Samples & History? \\ \hline\hline
ScreenSpot\cite{cheng2024seeclick}          & Web, PC, Mobile & Functional Grounding & ClickAcc                & 1,272           & ×       \\
Mind2Web\cite{deng2024mind2web}            & Web             & Navigation           & Ele.Acc, Op.F1, Step SR & 6,418           & \checkmark      \\
AMEX\cite{chai2024amex}                 & Mobile          & Navigation           & AMS                     & 5,284           & \checkmark      \\
GUI-Odyssey\cite{lu2024GUI-Odyssey}          & Mobile          & Navigation           & AMS                     & 29,426          & \checkmark      \\
AndroidControl-High\cite{li2024androidcontrol}  & Mobile          & Navigation           & Step Accuracy      & 8,444           & \checkmark      \\
AndroidControl-Low\cite{li2024androidcontrol}   & Mobile          & Functional Grounding & Step Accuracy      & 8,444           & ×       \\
GUIAct-Multi\cite{chen2024guicourse}         & Web             & Navigation           & StepSR                  & 1,065           & \checkmark      \\
GUIAct-Single\cite{chen2024guicourse}        & Web             & Functional Grounding & StepSR                  & 1,410           & ×       \\ \hline
\end{tabular}
\end{center}
\caption{Statistics of GUI benchmarks we include in this paper.}
\label{tab:statistics-benchmarks}
\end{table*}

\subsection{Metrics}
\label{appendix:metrics}

We use the metrics proposed in the corresponding benchmarks as shown in \cref{tab:statistics-benchmarks}. Although they have different names, these metrics are similar: GUI grounding tasks consistently measure the hit rate of predicted bounding boxes, while GUI navigation tasks focus on single-step accuracy.

\textbf{Click Accuracy}. The proportion of test samples where the predicted location falls in the ground truth element bounding box.

\textbf{Element Accuracy (Ele.Acc)}. Comparing the selected element with all acceptable elements. For vision-based methods, it is the same as Click Accuracy.

\textbf{Operation F1 (Op.F1)}. Token-level F1 score for the predicted operation.

\textbf{Step Success Rate (Step SR) \& Step Accuracy}. The proportion of successful steps. A step is regarded as successful only if both the selected element and the predicted operation are correct.

\textbf{Action Matching Score (AMS)}. The proportion of predicted actions that match the ground-truth actions. Two actions can match if their action types are equal. For dual-point taps, they are considered equal if they fall within a 14\% screen distance from each other. Alternatively, if the tap actions occur within the same detected bounding boxes, where the bounding boxes are augmented to 240\% of their total size, they are considered equal. Finally, two dual-point scrolls are considered equal if they have the same primary scroll axis (vertical or horizontal).

\subsection{Grounding Abilities for GUI Visual Appearances}
\label{appendix:gui-grounding-abilities}

To evaluate the foundational ability of SpiritSight to ground visual appearance, we construct a small benchmark for text2bbox task. We random select a small number of URLs from those mentioned in \cref{sec:level-1}. These URLs are not used in constructing the pre-training data. Following the method described in \cref{sec:level-1}, we construct a test set with 3,700 text2bbox pairs. We adopt the hit rate as metric, defined to be the proportion of test samples where the model predicted location falls within the ground-truth bounding boxes. Ultimately, SpiritSight-2B achieves a \textbf{96.1\%} hit rate on this benchmark, demonstrating its strong capability in fundamental grounding tasks. \cref{fig:vis-text2bbox} shows the visualization of the predicted bounding boxes from SpiritSight-2B.

\begin{table*}[t]
\begin{center}
\small
\begin{tabular}{ccccccc}
\hline
       & Dataset                   & Platform & \# Samples & \# Tokens & \# Elements & \# Screenshots \\ \hline\hline
\multirow{5}{*}{DocVQA} & Ureader-Instruction          & General & 489,150 & 23,356,600  & 489,150   & 118,355    \\
       & GUIChat                   & Web      & 50,832         & 14,789,523     & 50,832      & 17,979         \\
       & WebSRC                    & Web      & 23,742         & 162,177        & 23,742      & 5,462          \\
       & ScreenQA                  & Mobile   & 11,781         & 29,897         & 11,781      & 66,261\textsuperscript{\textdagger}       \\
       & RICO screen-captioning    & Mobile   & 15,743         & 121,854        & 15,743      & 66,261\textsuperscript{\textdagger}       \\ \hline
\multirow{5}{*}{Level1} & Web bbox2dom     & Web     & 862,505 & 281,389,127 & 862,505   & 755,499\textsuperscript{*} \\
       & Web text2bbox & Web      & 736,826        & 202,821,759    & 11,959,607  & 755,499\textsuperscript{*}     \\
       & Web bbox2text & Web      & 267,955        & 48,760,131     & 5,118,237   & 755,499\textsuperscript{*}     \\
       & AITW text2bbox            & Mobile   & 1,058,638
      & 309,594,824    & 27,993,054  & 1,276,752\textsuperscript{\textdaggerdbl}   \\
       & RICO widget-captioning    & Mobile   & 28,818         & 2,059,165      & 179,144     & 66,261\textsuperscript{\textdagger}       \\ \hline
\multirow{3}{*}{Level2} & Web function2bbox & Web     & 906,087 & 156,273,895 & 9,710,488 & 755,499\textsuperscript{*} \\
       & AITW function2bbox        & Mobile   & 620,736        & 18,542,781     & 620,736     & 1,276,752\textsuperscript{\textdaggerdbl}   \\
       & RICO-SCA                  & Mobile   & 18,148         & 2,485,239      & 145,517     & 66,261\textsuperscript{\textdagger}       \\ \hline
Level3 & AITW w/ CoT               & Mobile   & 639,535        & 53,039,652     & 639,535     & 1,276,752\textsuperscript{\textdaggerdbl}   \\ \hline\hline
Total  &                           &          & 5,730,496      & 1,113,426,624  & 57,820,071  & 2,240,308      \\ \hline
\end{tabular}
\end{center}
\caption{Statistics of our GUI-Lasagne, a GUI continual pre-training dataset for our SpiritSight-Base Model. In the '\# Screenshots' column, several datasets share the same suite of screenshot images, so numbers marked with the same superscript notation are counted only once.}
\label{tab:statistics-datasets}
\end{table*}

\begin{figure*}[t]
  \centering
  \begin{subfigure}{0.3\textwidth}
    \centering
    \raisebox{0mm}{
    \includegraphics[width=\linewidth]{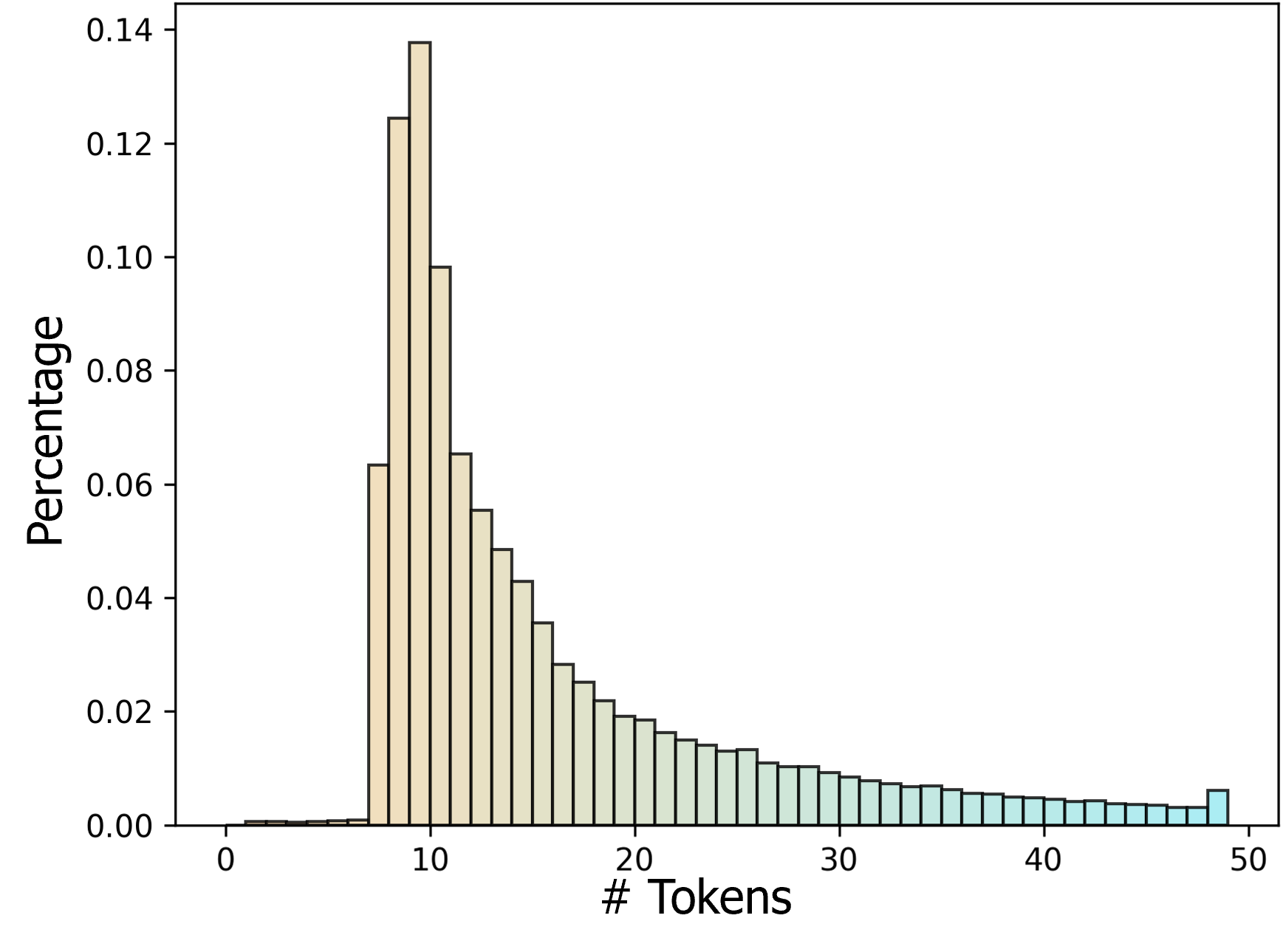}}
    \caption{GUI-Lasagne Level 1.}
    \label{fig:token_num_1}
  \end{subfigure}%
  \hfill %
  \begin{subfigure}{0.3\textwidth}
    \centering
    \includegraphics[width=\linewidth]{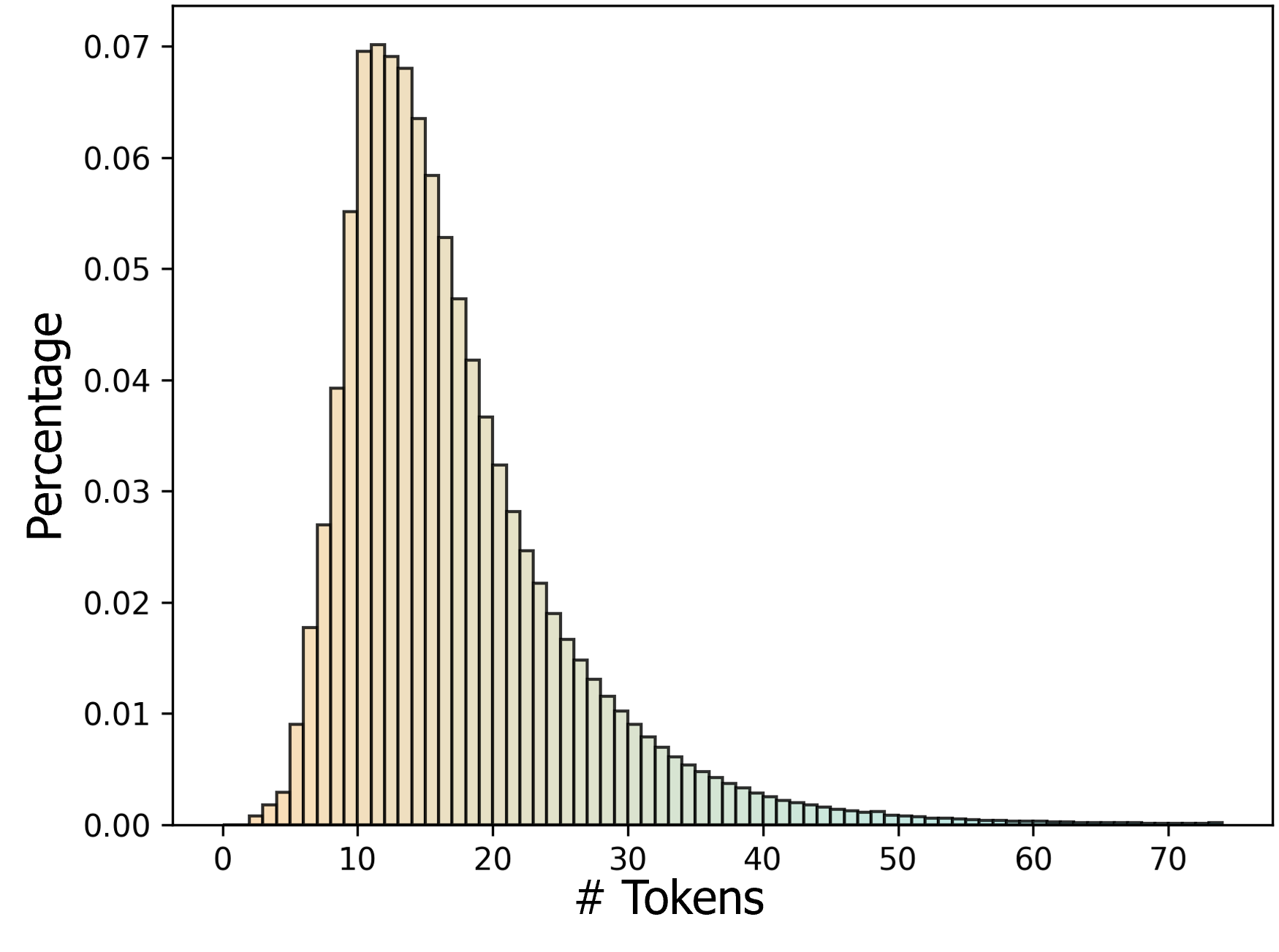}
    \caption{GUI-Lasagne Level 2.}
    \label{fig:token_num_2}
  \end{subfigure}
  \hfill %
  \begin{subfigure}{0.3\textwidth}
    \centering
    \includegraphics[width=\linewidth]{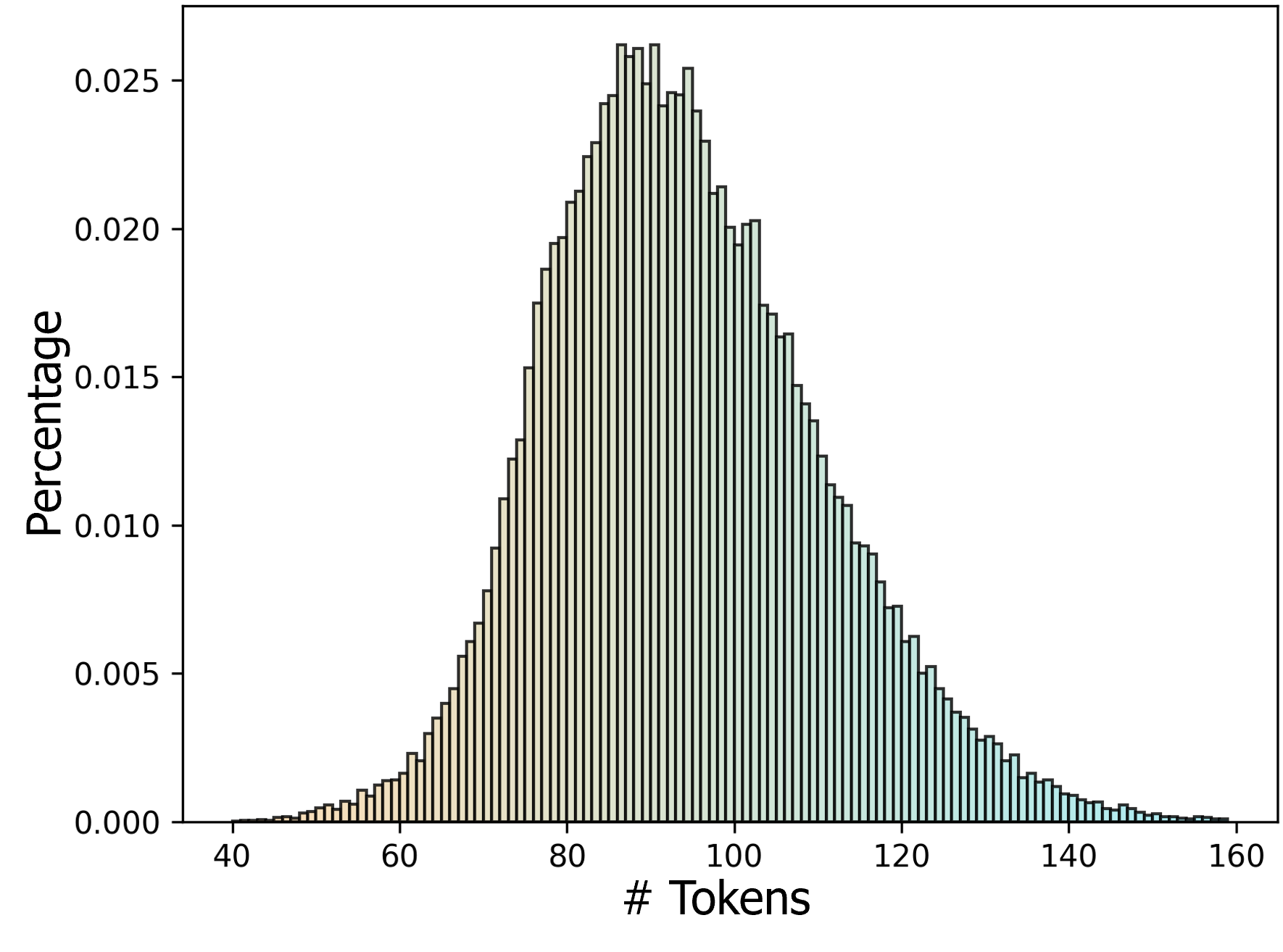}
    \caption{GUI-Lasagne Level 3.}
    \label{fig:token_num_3}
  \end{subfigure}
  \caption{The distribution of token numbers of our GUI-Lasagne dataset for GUI continual pretraining.}
  \label{fig:token_nums}
\end{figure*}

\begin{figure*}[t]
\label{fig:bbox2dom}
  \centering
    \includegraphics[width=0.99\linewidth]{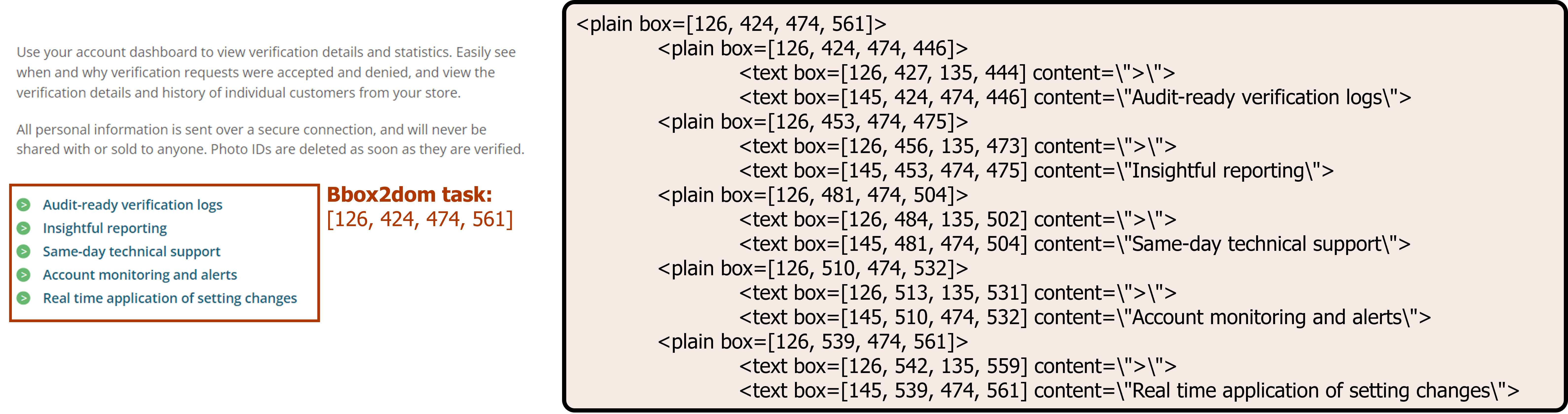}
    \caption{An example of the Bbox2dom task. Left shows a given bounding box on a web page, right shows its corresponding simplified DOM structure.}
\end{figure*}

\section{Data Collection}
\label{appendix:data-collection}

In this section, we present a cost-effective data collection strategy designed to construct a multi-level, large-scale and high-quality GUI dataset, called \textbf{GUI-Lasagne}. This dataset helps equip our models with robust abilities in GUI understanding, grounding, and navigation. The statistics of GUI-Lasagne are shown in \cref{tab:statistics-datasets} and \cref{fig:token_nums}.

\subsection{Level One: Visual-Text Alignment}
\label{appendix:data-collection-l1}

We collected website URLs from two sources: the CommonCrawl~\cite{cc2024Common-Crawl} dataset and website rankings. We used the URLs from website rankings as a supplement to CommonCrawl due to its compromised quality, which includes a large proportion of blank pages, sparse-texted pages, and dead links. We then developed a data collection tool using playwright library to get real-world web data from the collected URLs.

For each URL, we navigate to the webpage and start data collection only after the webpage has fully loaded. We collect both the webpage screenshots and the corresponding DOM tree according to a carefully designed scheme. First, we perform grid sampling on the screen with a step size of 8 pixels. Then, we mark the element objects corresponding to the sampled points. Finally, we apply an HTML pruning algorithm to simplify the HTML code by retaining all the marked elements and their parent nodes. This process excludes elements that are small in size or invisible on the screen. Additionally, we label all the clickable elements by checking their pointer property and registered events. The resulting DOM trees are used to construct the bbox2dom pairs, while the element objects are utilized to create the text2bbox and bbox2text pairs.

After collecting data from the current website, we acquire new pages using two methods: scrolling down or clicking on an element, with these choices being randomly sampled. If clicking on an element is chosen, the target element is also randomly sampled from all clickable elements. We collect a maximum of 30 pages for each URL. We repeat the above mentioned process to achieve an automated data collection. Ultimately, we collect 755K webpage screenshots along with their DOM trees, where English samples account for 3/4 of the data and Chinese samples account for 1/4.

\subsection{Level Two: Visual-Text Alignment}
\label{appendix:data-collection-l2}

We leverage InternVL's image understanding capabilities to collect function grounding data. Specifically, we divide each screenshot into a 3x3 grid and describe the approximate location of the target element in text format (\eg in the top-left corner of the image). Additionally, We place a bounding box around the target element in the screenshot to precisely specify its location. To prevent color confusion, we dynamically determine the color of the bounding box. First, we analyze the color data around the target element, then select the most visually prominent color among red, green, and blue as the color of bounding box. By providing InternVL2-26B with the screenshot, the element's text content or icon caption, and the location description, we prompt it to generate the function of the target element. Additionally, we utilize InternLM2.5-20B to enhance the quality and diversity of the generated function descriptions. The two prompts are shown in \cref{appendix:prompt-l2}.

A validation is performed by two experienced human annotators. Specifically, we randomly sampled 100 images from the collected data to create a human evaluation set, with the functional description of all labeled element. The annotators are asked to determine whether the functional description is correct. A functional description is considered acceptable if the corresponding element can be uniquely identified in the screenshot based on the description. We calculate the proportion of acceptable functional descriptions out of the total descriptions. Ultimately, the human evaluation achieves an acceptance rate of 90.9\%, indicating the effectiveness of our data synthesis strategy. We show some evaluation examples in \cref{fig:example-level2-3}.

\subsection{Level Three: Visual GUI Navigation}
\label{appendix:data-collection-l3}

We utilize the public available AitW~\cite{rawles2024AndroidintheWild} dataset to construct our GUI navigation training data. AitW is a large-scale mobile navigation dataset where each screenshot is labeled with the corresponding goal, the current step, \emph{etc.} We select the all general, install and web-shopping sets and 1M samples of google-apps set as the source data. We discard the single set as the screenshots are duplicated with others. However, AitW involves a certain amount of incorrectly labeled samples as mentioned by AitZ~\cite{zhang2024AndroidintheZoo} and AMEX~\cite{chai2024amex}.

We clean the AitW~\cite{rawles2024AndroidintheWild} dataset with GPT-4o and adopt Chain-of-Thought (CoT)~\cite{wei2022CoT} to make the judgment more accurate. Specifically, for non-final steps, we prompt GPT-4o with the task description, the current action annotation, two screenshots at the current and the next steps, respectively. GPT-4o is then instructed to first summarize the two screenshots and identify the differences between them, then describe the current step based on these differences, and finally assess the reasonableness of the current action annotation. We filter out steps identified as unreasonable by GPT-4o. For the final step, we prompt GPT-4o with the task description and the current screenshot. It is then instructed to summarize the screenshot and determine whether the task was successfully completed. We filter out the final steps considered successfully completed. Note that we only discard the steps that do not meet the requirements, and do not discard the entire trajectories. The prompts for GPT-4o are shown in \cref{appendix:prompt-l3}. The collected data examples are shown in \cref{fig:example-level2-3}. Ultimately, we obtain 0.63M CoT-style GUI navigation training samples from 1.48M source samples after cleaning.

We also perform a validation for level-3 data by two experienced human annotators. Specifically, We randomly sampled 100 steps that considered as reasonable (the Cleaned Set) and 100 discarded steps by GPT-4o (the Discarded Set). For each step sample, the annotators are provided with the screenshots, the overall task description, and the validity judgments generated by GPT-4o. Then they are asked to determine whether the results of GPT-4o is correct. We report the true positive rate (TPR) for the Cleaned Set and the true negative rate (TNR) for the Discarded Set. The Cleaned Set achieved a TPR of 93.7\%, indicating the reliability of our data cleaning procedure. The Discarded Set achieved a TNR of 76.3\%. Though the result is not as high, it is unrelated to the quality of our dataset. In the future, we will explore a more efficient data cleaning method to improve the TNR while keeping the TPR approximately unchanged.

\section{Training Data Format}
\label{appendix:training-data-format}

We constructed a large scale dataset for GUI continual pre-training, including text2bbox, bbox2text, bbox2dom, and function2bbox tasks. To make sufficient use of the context length of the model, we pack multiple data pairs in each training sample for text2bbox, bbox2text and function2bbox tasks, and select the box that includes as many elements as possible for bbox2dom task. We use the center point, width and height to represent a bounding box. It is worth noting that, aside from the function2bbox task, we add an additional block index to each bounding box, which is derived from our proposed UBP method. For function2bbox task, we use the original global coordinate system as the bounding boxes are too large to be considered a point and grounding is not the main focus of this task. Additionally, we normalize all coordinate values between 0 and 999 and round them to the nearest integer. Below are the training data templates for each task. Notably, the prompt is randomly selected from a pool during data construction. See \cref{appendix:prompt} for details of the prompt pool.

    \begin{tcolorbox}[
    colframe=black,
    colback=blue!10!white]
    \textbf{\textit{Data Format for text2bbox Task}}\\
    \textit{user:}\\
    $<$image$>$\\
    1.\{text 1\}\\
    2.\{text 2\}\\
    3.\{text 3\}\\ ...\\
    Provide the bounding boxes of each given text in a list format.\\
    \\
    \textit{assistant:}\\
    1.\{[block-index, cx, cy, w, h]\}\\
    2.\{[block-index, cx, cy, w, h]\}\\
    3.\{[block-index, cx, cy, w, h]\}\\ ...
    \end{tcolorbox}

    \begin{tcolorbox}[
    colframe=black,
    colback=blue!10!white]
    \textbf{\textit{Data Format for bbox2text Task}}\\
    \textit{user:}\\
    $<$image$>$\\
    1.\{[block-index, cx, cy, w, h]\}\\
    2.\{[block-index, cx, cy, w, h]\}\\
    3.\{[block-index, cx, cy, w, h]\}\\ ...\\
    Provide the text content of each given bounding box in a list format.\\
    \\
    \textit{assistant:}\\
    1.\{text 1\}\\
    2.\{text 2\}\\
    3.\{text 3\}\\ ...
    \end{tcolorbox}

    \begin{tcolorbox}[
    colframe=black,
    colback=blue!10!white]
    \textbf{\textit{Data Format for bbox2dom Task}}\\
    \textit{user:}\\
    $<$image$>$\\
    I'd like some information about the specific region [cx, cy, w, h] in the image. \\
    \\
    \textit{assistant:}\\
    \{DOM\_Tree\}
    \end{tcolorbox}

    \begin{tcolorbox}[
        colframe=black,
        colback=blue!10!white]
    \textbf{\textit{Data Format for function2bbox Task}}\\
    \textit{user:}\\
    $<$image$>$\\
    1.\{function description 1\}\\
    2.\{function description 2\}\\
    3.\{function description 3\}\\ ...\\
    In this image from a webpage, find out where to click for a certain need and provide bbox coordinates in a list format.\\
    \\
    \textit{assistant}\\
    1.\{[block-index, cx, cy, w, h]\}\\
    2.\{[block-index, cx, cy, w, h]\}\\
    3.\{[block-index, cx, cy, w, h]\}\\ ...
    \end{tcolorbox}

\section{Prompt Templates}
\label{B-prompt-templates}

\subsection{Evaluation Inference}
\label{appendix:prompt}

\begin{tcolorbox}[
    colframe=black,
    colback=yellow!10!white]
\textbf{\textit{Prompt for Evaluation Inference}}\\
\#\# Task: \{task\} \\
\#\# History Actions: \\
\{history\} \\
\#\# Action Space \\
\{Action Space\} \\
\#\# Requirements: Please infer the next action according to the Task and History Actions. \\
Return with Action Code. The Action Code should follow the definition in the Action Space.
\end{tcolorbox}

\subsection{Level-two Function Generation}
\label{appendix:prompt-l2}

\begin{tcolorbox}[
    colframe=black,
    colback=yellow!10!white]
\textbf{\textit{Prompt for Level-two Function Generation}}\\
Please infer the purpose of the operation "click on the '\{text\}' on the \{region\} of the webpage" based on the webpage.\\
Please deliver the purpose specifically and clearly, which points to the certain item.\\
Its direct context includes the following information: \{context\_text\}.\\
Please make the answer only in English.\\
Let's think step by step.\\
Your final answer should be in a new line and included in double quotation like:\\
The purpose is "xxx".
\end{tcolorbox}

\begin{tcolorbox}[
    colframe=black,
    colback=yellow!10!white]
\textbf{\textit{Prompt for Level-two Function Augmentation}}\\
Can you rewrite the original purpose "\{purpose\}" into a short phrase?\\
Here are some examples:\\
\{Few-shot example 1\}\\
\{Few-shot example 2\}\\
\{Few-shot example 3\}\\
Output only the refined purpose, start with 'to', without any explanation.
\end{tcolorbox}

\subsection{Level-three Data Processing}
\label{appendix:prompt-l3}

\begin{tcolorbox}[
    colframe=black,
    colback=yellow!10!white]
\textbf{\textit{System Prompt for Level-three Data Processing}}\\
You are a mobile operation assistant, the main goal is to help identify whether the mobile navigation operation is correct.
\end{tcolorbox}

\begin{tcolorbox}[
    colframe=black,
    colback=yellow!10!white]
\textbf{\textit{Prompt for Level-three Middle Step Data Processing}}\\
Task: \{task\}\\
Action History: \{history\}\\
The Current Action: \{action\}\\
You are completing a mobile task and now in step \{step\_idx\}. Picture 1 shows the current screen with action demonstration and picture 2 shows the screen after performing The Current Action on picture 1. You are also given the Action History before the Current Action.\\
Return:\\
1. Summarize picture 1 about its main content and its functionality. Also describe the changes that have occurred in Figure 2 compared to Figure 1. Describe them with necessary details, but not too long.\\
2. Based on the changes between Figure 1 and Figure 2, estimate the function of the Current Action. Return with format of "The function of the Current Action: xxx"\\
3. Analyze the rationality of the Current Action based on the Task. Return only the reason.\\
4. Return the final answer of the rationality of the Current Action with just 'True' or 'False'.\\
5. Analyze if the Task is successfully completed. Return only the reason.\\
6. Return the final answer of the complementarity of the Task with just 'True' or 'False'.
\end{tcolorbox}

\begin{tcolorbox}[
    colframe=black,
    colback=yellow!10!white]
\textbf{\textit{Prompt for Level-three Last Step Data Processing}}\\
Task: \{task\}\\
Action History: \{history\}\\
You have just completed a mobile task with a series of actions listed in Action History. The picture shows the final screen of the mobile.\\
Return:\\
1. Summarize the picture about its main content and its functionality. Describe it with necessary details, but not too long.\\
2. Analyze if the task is successfully completed from the perspectives of success and completion separately.\\
3. Return the final answer of the analysis with just 'True' or 'False'.
\end{tcolorbox}

\begin{figure*}[t]
  \centering
    \includegraphics[width=0.98\linewidth]{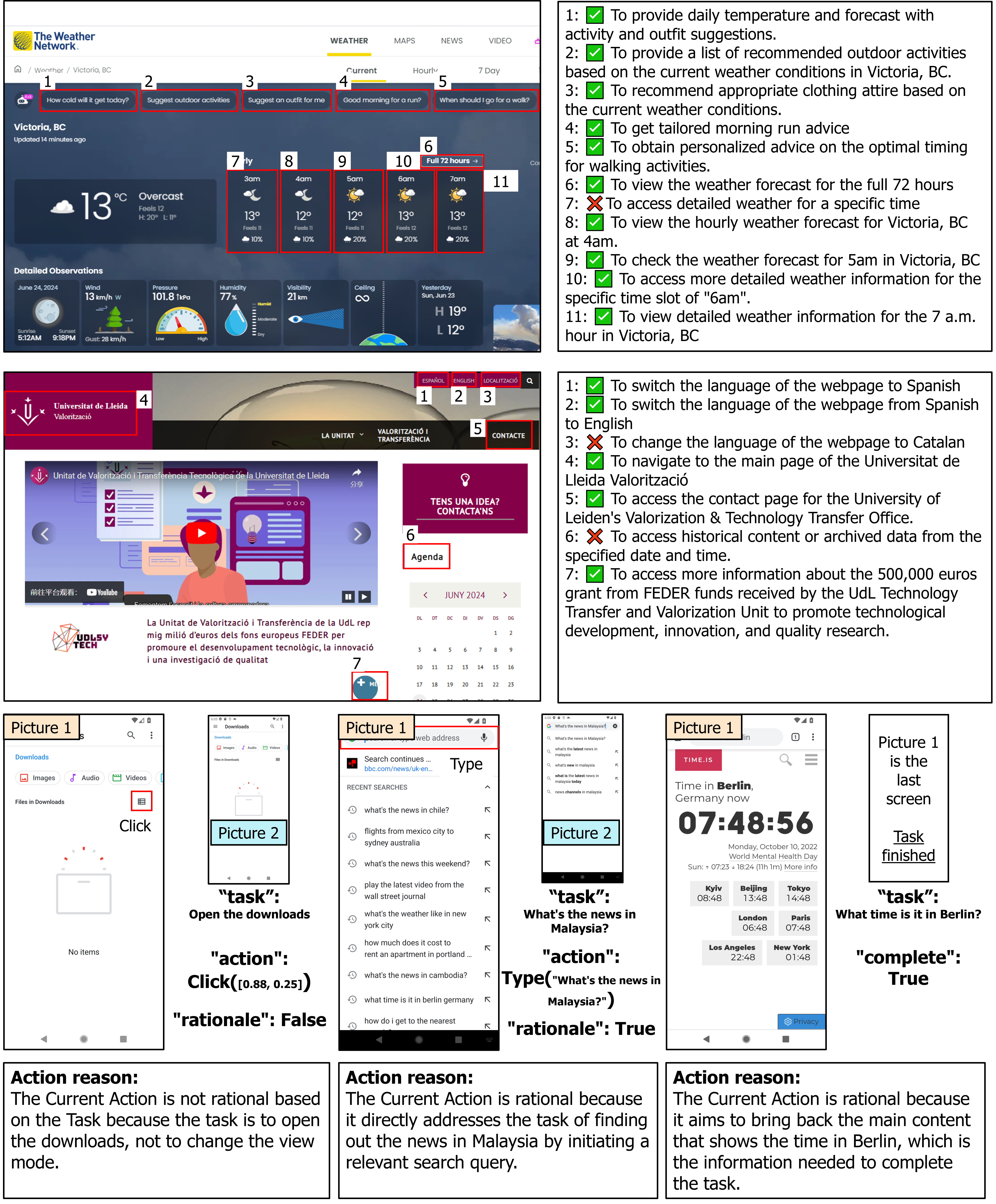}
    \caption{The examples of our collected GUI function and navigation data. The upper two screenshots show the functional annotation generated by InternVL2 and InternLM2.5. The lower three samples show the judgment results and reasons provided by GPT-4o.}
  \label{fig:example-level2-3}
\end{figure*}

\fi

\end{document}